\documentclass{article}

\usepackage[preprint]{neurips_2025}

\usepackage[utf8]{inputenc} 
\usepackage[T1]{fontenc}    
\usepackage{url}            
\usepackage{booktabs}       
\usepackage{amsfonts}       
\usepackage{nicefrac}       
\usepackage{microtype}      
\usepackage{amsmath,amsfonts,bm}

\def\eqref#1{equation~\ref{#1}}

\def\1{\bm{1}}

\DeclareMathAlphabet{\mathsfit}{\encodingdefault}{\sfdefault}{m}{sl}
\SetMathAlphabet{\mathsfit}{bold}{\encodingdefault}{\sfdefault}{bx}{n}

\newcommand{\normltwo}{L^2}

\usepackage[utf8]{inputenc} 
\usepackage[T1]{fontenc}    
\usepackage{url}            
\usepackage{booktabs}       
\usepackage{amsfonts}       
\usepackage{nicefrac}       
\usepackage{microtype}      
\usepackage[dvipsnames]{xcolor}         
\usepackage{colortbl}
\usepackage[ruled,linesnumbered]{algorithm2e}
\usepackage{graphicx}
\usepackage{svg} 
\usepackage{enumitem}
\usepackage{pifont}   
\usepackage{subcaption}
\usepackage{mdframed}
\usepackage[normalem]{ulem}
\usepackage[most]{tcolorbox}   
\usepackage{tikz}
\usepackage{float}
\usetikzlibrary{backgrounds}
\usetikzlibrary{arrows,shapes}
\usetikzlibrary{tikzmark}
\usetikzlibrary{calc}
\usepackage{fontawesome}
\usepackage{nccmath}
\usepackage{wrapfig}
\usepackage{comment}
\usepackage{amsmath}
\usepackage{amssymb}
\usepackage{bbm}
\usepackage{mathtools}
\usepackage{amsthm}
\usepackage{diagbox}
\usepackage{multirow} 
\usepackage{bm}
\usepackage{float}
\usepackage[backref=page]{hyperref}
\usepackage[capitalize,nameinlink]{cleveref}[0.19]

\crefname{definition}{Definition}{Definitions}
\crefname{assumption}{Assumption}{Assumptions}
\crefname{theorem}{Theorem}{Theorems}
\crefname{remark}{Remark}{Remarks}
\crefname{lemma}{Lemma}{Lemmas}
\crefname{corollary}{Corollary}{Corollaries}
\crefname{proposition}{Proposition}{Propositions}
\crefname{section}{Section}{Sections}
\crefname{example}{Example}{Examples}
\crefname{table}{Table}{Tables}
\crefname{problem}{Problem}{Problems}
\crefname{algorithm}{Algorithm}{Algorithms}
\crefname{figure}{Figure}{Figures}
\crefname{property}{Property}{Properties}
\crefformat{equation}{#2Equation~(#1)#3}
\crefformat{inequality}{#2Inequality~{}(#1){}#3}
\crefrangeformat{inequality}{#3Inequality~(#1)#4--#5(#2)#6}
\Crefformat{section}{#2§#1#3}
\Crefformat{subsection}{#2§#1#3}
\Crefformat{page}{#2Page~#1#3}

\definecolor{darkgrey}{rgb}{0.53,0.53,0.53}
\definecolor{mygrey}{rgb}{0.85,0.85,0.85}
\definecolor{darkblue}{rgb}{0,0.08,0.45}

\definecolor{darkred}{HTML}{C00000}
\definecolor{lightblue}{HTML}{F9FEFE}
\definecolor{fancyblue}{HTML}{4771E3}
\hypersetup{ %
pdftitle={},
pdfauthor={},
pdfsubject={},
pdfkeywords={},
pdfborder=0 0 0,
pdfpagemode=UseNone,
colorlinks=true,
linkcolor=darkred,
citecolor=darkblue,
filecolor=darkblue,
urlcolor=darkblue}

\newif\iffoo

\colorlet{LightBlue}{blue!39!white} 
\colorlet{DarkBlue}{blue!70!black} 
\colorlet{VeryLightBlue}{blue!30!white} 
\colorlet{LightRed}{red!35!white} 

\usepackage[textsize=tiny]{todonotes}

\title{Soft Weighted Machine Unlearning}

\author{%
  Xinbao Qiao$^{1}$ 
  \And Ningning Ding$^{2}$
  \And Yushi Cheng$^{1}$ 
  \And Meng Zhang$^{1}$ 
  \AND \vspace*{-5mm}\\
  ${}^{1}$Zhejiang University,
  ${}^2$The Hong Kong University of Science and Technology (Guangzhou)\\
  \texttt{xinbaoqiao@zju.edu.cn}\\
}

\makeatletter
\pretocmd{\tableofcontents}{\hypersetup{linkcolor=black}}{}{}
\apptocmd{\tableofcontents}{\hypersetup{linkcolor=darkred}}{}{}
\makeatother

\begin{document}

\maketitle

\begin{abstract}
Machine unlearning, as a post-hoc processing technique, has gained widespread adoption in addressing challenges like bias mitigation and robustness enhancement, colloquially, machine unlearning for fairness and robustness. However, existing non-privacy unlearning-based solutions persist in using binary data removal framework designed for privacy-driven motivation, leading to significant information loss, a phenomenon known as “over-unlearning”. While over-unlearning has been largely described in many studies as primarily causing utility degradation, we investigate its fundamental causes and provide deeper insights in this work through counterfactual leave-one-out analysis.
In this paper, we introduce a weighted influence function that assigns tailored weights to each sample by solving a convex quadratic programming problem analytically. Building on this, we propose a soft-weighted framework enabling fine-grained model adjustments to address the over-unlearning challenge. We demonstrate that the proposed soft-weighted scheme is versatile and can be seamlessly integrated into most existing unlearning algorithms.
Extensive experiments show that in fairness- and robustness-driven tasks, the soft-weighted scheme significantly outperforms hard-weighted schemes in fairness/robustness metrics and alleviates the decline in utility metric, thereby enhancing machine unlearning algorithm as an effective correction solution.
\end{abstract}

\section{Introduction}
\label{SEC-1}
Modern machine learning (ML) models benefit greatly from the quantity and quality of the training data they are built upon. Depending on the type of the trained model being used, the impact
of training samples can be either beneficial or detrimental.  As a recent advancement, machine unlearning, originally conceived as a privacy-preserving mechanism to comply with data protection regulations' “the right to be forgotten” by allowing users to remove their personal data from models, has significantly broadened its scope. Beyond its privacy-oriented motivation, machine unlearning, as a post-hoc technique, has recently addressed broader practical concerns in trained models through efficient data removal, e.g., correcting bias \cite{chen2024fast,oesterling2024fair} and mitigating the detrimental effects \cite{liu2022backdoor,zhang2023recommendation,li2024delta,kurmanji2024machine}. These applications provide a fast way to adapt and edit a trained model without the prohibitively expensive process of retraining from scratch, catalyzing a paradigm shift in machine unlearning methodologies to address critical challenges beyond privacy concerns.

However, influenced by the inertia of prior research rooted in privacy-centric considerations, these traditional methods solving non-privacy challenges operate under a binary framework: data is to remove or not to remove,  which we refer to as hard-weighted unlearning framework in this paper, characterized by the complete elimination of undesired data influences. This framework, while suitable for stringent privacy requirements, presents significant limitations when addressing more complex non-privacy-oriented challenges in modern ML systems, where the objective has transformed from regulatory-mandated data deletion to tasks such as enhancing model fairness, adversarial robustness, and generalization capabilities.

Specifically, the hard-weighted unlearning framework introduces several critical challenges: potential overcorrection, significant information loss, and compromised model generalization, collectively
defined as \textbf{over-unlearning} by numerous studies \cite{DBLP:conf/ndss/Hu0CZ00ZX24,DBLP:conf/ijcai/ChenZ0024}. The binary nature of hard-weighted decisions can lead to suboptimal outcomes, particularly when dealing with nuanced data distributions or complex objectives. We illustrate this concretely as evidence in \cref{Fig: SEC-1.1}, where we trained a linear model on Adult dataset \cite{adult_2} (See \cref{Appendix-Visualizing the Correlations} for the results of other datasets) and analyzed the performance of leave-one-out models obtained by removing each sample individually. Specifically, we evaluated changes in the following metrics as the differences between their post-removal and pre-removal values: fairness, quantified by Demographic Parity \cite{DBLP:conf/innovations/DworkHPRZ12}; adversarial robustness, assessed through the loss on perturbed datasets \cite{DBLP:conf/esann/MegyeriHJ19}; and generalization utility, determined by the loss on the test set. 
These results allowed us to uncover the underlying causes of over-unlearning:
\begin{wrapfigure}{r}{0.465\textwidth}
\vspace{-1em}
\includegraphics[width=\linewidth]{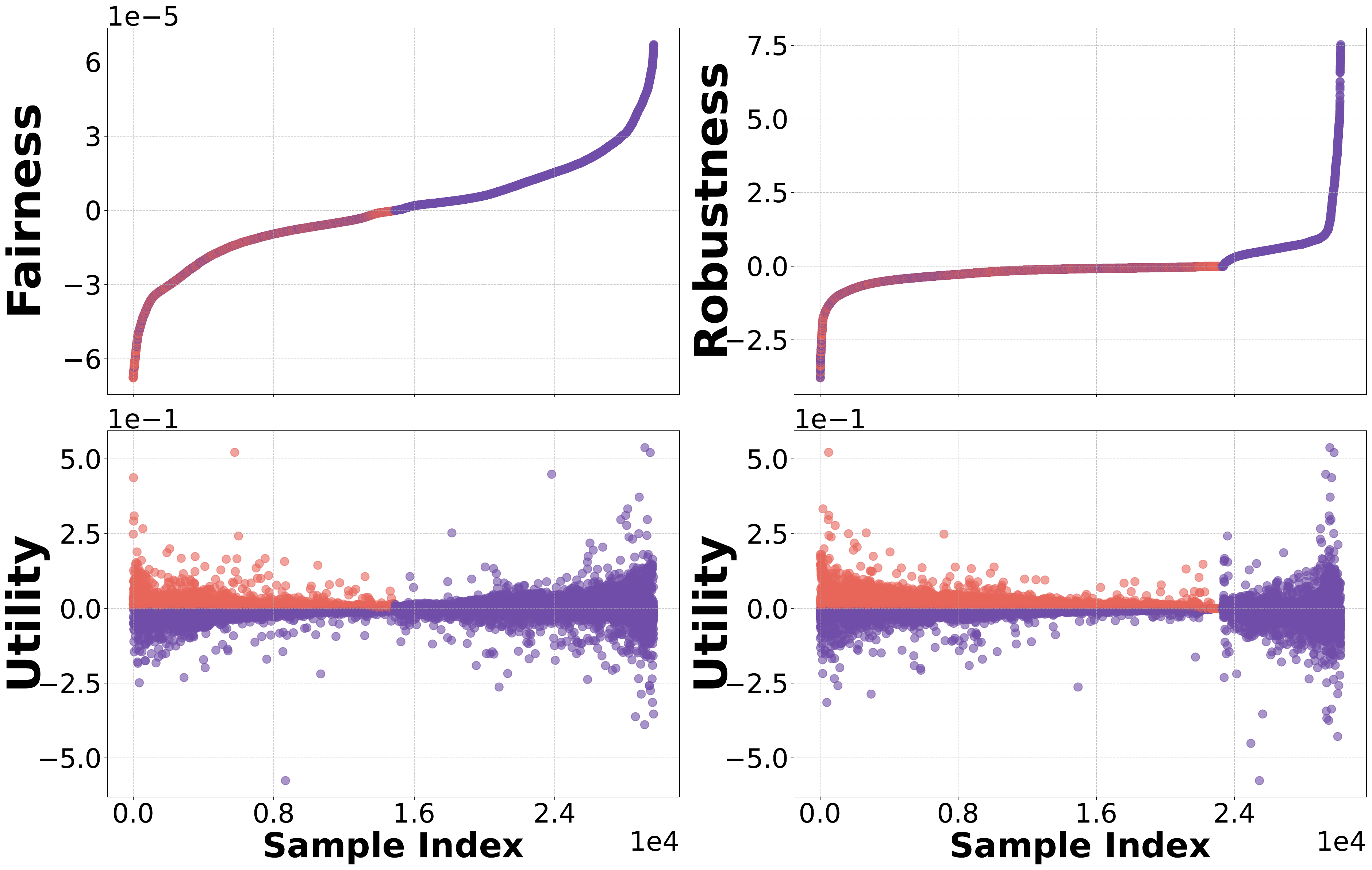}
\vspace{-1em}
\caption{\textbf{Actual Changes in Utility and Fairness/Robustness on Adult dataset} for each sample's leave-one-out model. \textbf{The X-axis represents} the sample indices. \textbf{The Y-axis for Fairness (Robustness)} displays changes in demographic parity (adversarial loss) on the test set, with negative values indicating improved fairness (robustness) and positive values indicating reduced fairness (robustness). \textbf{The Y-axis for Utility} shows changes in test loss, with negative values indicating improved utility. Scatter points marked in \textcolor[HTML]{E8655A}{\textbf{Red}} indicate sample indices where fairness or robustness improves, but utility (generalization) declines.
}
    \label{Fig: SEC-1.1}
\end{wrapfigure}
\textbf{\ding{182} Fairness/Robustness and utility are uncorrelated.} The overall Spearman correlation coefficients for fairness/robustness and utility across all samples are -0.11 and -0.16 respectively, meaning that improvements in the target task do not always translate to better utility. Similarly, the red-highlighted samples in \cref{Fig: SEC-1.1} indicate that removing the most detrimental samples does not lead to accuracy gains, highlighting the primary cause of over-unlearning.
\\
\textbf{\ding{183} Borderline forgetting samples are treated equivalently to highly detrimental samples by unlearning algorithm.}  The majority of forgetting samples are not the main contributors to model bias (vulnerability). However, in hard-weighted frameworks, such as gradient ascent algorithms \cite{DBLP:conf/nips/JiaLRYLLSL23}, the samples are treated uniformly in an attempt to remove the most biased (vulnerable) ones, which can lead to excessive unlearning of borderline samples. This may cause borderline samples to be flipped to unprivileged groups, resulting in opposite biases, or it may have an opposite effect on robustness.
\\
\textbf{\ding{184} The majority of detrimental samples are maintained in remaining dataset.}  
Approximately the top 50\% (75\%) of samples with values below 0 in \cref{Fig: SEC-1.1} exacerbate model bias and vulnerability. However, existing algorithms under the hard-weighted framework \cite{chen2024fast} for unlearning 20\% of the samples can only remove a limited number of samples and struggle to support further data removal. 

\begin{figure*}
    \centering
\includegraphics[width=0.95\linewidth]{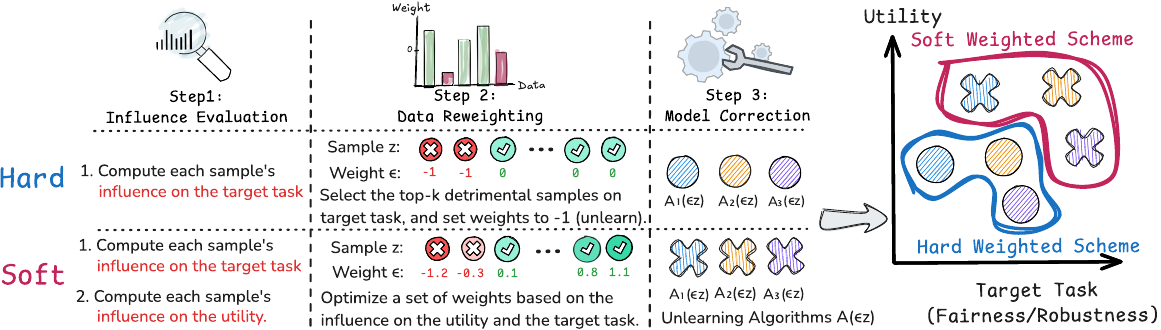}
    \caption{Illustration of difference of the proposed \textcolor[HTML]{c2255c}{\textbf{soft}} weighted vs. the \textcolor[HTML]{1971c2}{\textbf{hard}} weighted framework. }
    \label{Fig: SEC-1.2}
\vspace{-0.7em}
\end{figure*}

In this paper, we take the first step in addressing the challenge of over-unlearning when applying machine unlearning to other domains. To the best of our knowledge, our work is the first to uncover the root causes of over-unlearning and propose a framework to tackle this issue. \cref{Fig: SEC-1.2} illustrates our conceptual framework and highlights its differences from prior works \cite{chen2024fast}. We use influence functions as a tool, enabling the interchangeable use of various influence-based methods, and extend their applicability to a wider range of domains and scenarios, such as adversarial robustness. The key difference lies in our departure from the binary removal scheme inherited from privacy-driven motivations, instead adopting an optimization approach that allocates weights to each data. This smoother, softer approach empirically demonstrates enhanced performance on target tasks while improving utility. We summarize our main contributions as follows:
\begin{itemize}[leftmargin=0.3343cm, itemindent=0cm]
    \item[$\bullet$] We reveal the deeper causes of over-unlearning challenge from the perspective of counterfactual analysis in \cref{SEC-1}, offering insights for the development of machine unlearning.
    \item[$\bullet$] We introduce the weighted influence function in \cref{SEC-4-1}, a refined solution to address this challenge, with the weights through solving a convex quadratic programming problem in \cref{SEC-4-2}. We demonstrate that the soft weighted framework in \cref{SEC-4-3} can be integrated into most unlearning methods.
    \item[$\bullet$] We empirically show in \cref{SEC-5} that the proposed framework significantly boosts the performance of most existing algorithms in fairness/robustness tasks as well as utility, with only a few seconds of additional time overhead. 
\end{itemize}

\newpage

\section{Related Works}
\vspace{-0.4em}

\textbf{Machine Unlearning}, including recent cutting-edge methods such as \cite{DBLP:conf/nips/KurmanjiTHT23,DBLP:journals/corr/abs-2201-06640,DBLP:conf/emnlp/ChenY23}, is claimed to address challenges beyond its original privacy concerns, e.g., tackling issues like debiasing or enhancing robustness in well-trained models. These methods typically follow a paradigm where data to be forgotten is provided through deletion requests, after which the unlearning process is executed. These algorithms require prior knowledge to identify which data needs to be forgotten. \cite{chen2024fast,zhang2023recommendation}  thus advanced an ``Evaluation then Removal” framework, utilizing influence functions \cite{DBLP:conf/icml/KohL17} for model debiasing. By using influence functions, the framework can first estimate the subset of data most responsible for model bias or vulnerability, thereby resolving the challenge of identifying forgetting dataset and subsequently unlearning undesired data. Furthermore, despite existing work exploring the fairness and robustness of machine unlearning methods, e.g., \cite{DBLP:conf/aistats/OesterlingMCL24,DBLP:journals/corr/abs-2404-15760,tran2025fairness,tamirisa2024toward,10.1145/3627673.3679817,DBLP:conf/cikm/Sheng0G24,dige2024mitigating}, these approaches primarily focus on enhancing the fairness and robustness of unlearning algorithms themselves, rather than leveraging machine unlearning for fairness \cite{chen2024fast} and robustness \cite{DBLP:conf/aaai/HuangSGL025} tasks. 

\textbf{Fairness} and related ethical principles are crucial in ML research. Most methods for addressing unfairness rely on the concept of (un)privileged groups, which are disproportionately (less) likely to receive favorable outcomes. Fairness definitions in the literature focus on either group or individual fairness. Group fairness compares outcomes across groups but may harm within-group fairness, while individual fairness, such as counterfactual fairness which requires generating counterfactual samples, aims to ensure fairness across individuals \cite{DBLP:conf/fat/HutchinsonM19}.
As pointed out in \cite{DBLP:journals/csur/CatonH24}, fairness notions are often incompatible and have limitations, with no universal metric or guideline for measuring fairness \cite{DBLP:journals/bigdata/Chouldechova17,kleinberg2018algorithmic}. Our study does not compare different fairness definitions but instead focuses on succinctly quantifying fairness using group fairness metrics, including Demographic Parity (DP) \cite{DBLP:conf/innovations/DworkHPRZ12} and Equal Opportunity (EOP) \cite{DBLP:conf/nips/HardtPNS16}, which are widely adopted in ML contexts \cite{DBLP:conf/iclr/ChhabraLM024}.
 
\textbf{Robustness}, or in other words, the vulnerability of ML model predictions to minor sample perturbations \cite{DBLP:conf/cvpr/EykholtEF0RXPKS18}, is another key aspect of ML research. In this paper, we focus on the influence of data on robustness. A related work \cite{DBLP:journals/csur/XiongTSPR24} summarizes the effects of data on adversarial robustness and highlights how to select data to enhance robustness. Similar to \cite{DBLP:conf/iclr/ChhabraLM024}, we explore a white-box attack strategy to craft adversarial samples \cite{DBLP:conf/esann/MegyeriHJ19} targeting a linear model, which can be extended to methods such as FGSM \cite{DBLP:journals/corr/GoodfellowSS14} and PGD \cite{DBLP:conf/iclr/MadryMSTV18}. We quantify robustness as performance under adversarial attacks, referred to as perturbed accuracy, which is distinguished from utility known as standard test accuracy. 


\vspace{-0.5em}
\section{Preliminaries}
\label{SEC-3}
\vspace{-0.5em}

 Let $\ell(z;\theta)$ be a loss function for a given parameter $\theta$ over parameter space $\Theta$ and sample $z$ over instance space $ \mathcal{Z}$. The empirical risk (ER) minimizer on the training dataset $\mathcal{D} \!\! =\! \{z_i\!\!=\!\!(\mathbf{x}_i, y_i)\}_{i=1}^{n}$ is given by $\hat{\theta} \!=\! \arg \min _{\theta \in \Theta} \frac{1}{n}\! \sum_{i\!=\!1}^n \! \ell\left(z_i ; \theta \right) $.  For the ER that is twice-differentiable and strictly convex\footnote{The convexity makes the theoretical analysis of influence functions impossible in non-convex models, yet this does not invalidate the use of influence functions in practice. In non-convex scenarios, these strategies are widely adopted: (i) using a convex surrogate model on embeddings from the non-convex model \cite{DBLP:conf/icml/GuoGHM20,chen2024fast}, (ii) adding a damping factor to ensure a positive definite Hessian \cite{DBLP:conf/icml/ZhangDWL24}, and (iii) reweighting gradient updates instead of loss in SGD-trained models, thereby avoiding the inversion of the Hessian \cite{qiao2024efficient}.} in parameter space $\Theta$, we slightly perturb the sample $z_j$ by reweighting it with weight $\epsilon_j \in  \mathbb{R}$,
 \vspace{-0.5em}
\begin{equation}
\label{Eq: EMR}
    \hat{\theta}(z_j ;\epsilon_j)\!=\!\arg \min _{\theta \in \Theta} \frac{1}{n} \! \sum_{i=1}^n \! \left(\ell\left(z_i; \theta\right) \!+\! \epsilon_j \ell\left(z_j ; \theta\right) \right).
\end{equation}
Let $\epsilon_j \!=\!-1$ give $ \hat{\theta}(z_j; -\!1) $, the ER minimizer trained without sample $ z_j $, and clearly, $ \hat{\theta} \!\!=\!\! \hat{\theta}(z_j; 0) $. Thus, using influence function \cite{DBLP:conf/icml/KohL17} can efficiently capture model change through closed-form update:
\vspace{-1em}
\begin{equation}
\label{Eq: Influence on Parameter}
\hat{\theta}(z_j ;-1)-\hat{\theta}(z_j ;0) \approx  \frac{1}{n} \mathbf{H}_{\hat{\theta}}^{-1} \nabla_{{\theta}} \ell(z_j ; \hat{\theta}),
\end{equation}

\vspace{-1.1em}
where $\mathbf{H}_{\hat{\theta}} \stackrel{\text { def }}{=} \frac{1}{n} \sum_{i-1}^n \nabla_\theta^2 \ell (z_i, \hat{\theta})$ is the Hessian matrix. See more details in \cref{APPENDIX-Technique Details}.
For a function $ f $ of interest, e.g., utility (generalization), fairness or robustness metrics, the actual change of function $ f $ is expressed as $ \mathcal{I}^*(z_j; \epsilon) = f(\hat{\theta}(z_j; \epsilon)) - f(\hat{\theta}) $, which can be efficiently estimated by:

\textbf{Utility:} $\mathcal{I}_{\text{util}}(z_j ;-1)\!=  \!\! \sum_{z \in \mathcal{T}}\! \!\nabla_{{\theta}} \ell( z ; \hat{\theta})^{\!\top}\! \mathbf{H}_{\hat{\theta}}^{-\!1} \nabla_{{\theta}} \ell(z_j; \hat{\theta})$, which reflects the loss change in the validation set $\mathcal{T}$, where a negative value indicates better generalization in model trained without $z_j$. 

\textbf{Fairness:}
  $\mathcal{I}_{\text {fair}}(z_j;-1 )=  
 \nabla_{{\theta}} f_{\text{fair}}(\mathcal{T} ; \hat{\theta})^{\top} \mathbf{H}_{\hat{\theta}}^{-1} \nabla_{{\theta}} \ell(z_j; \hat{\theta})$ .
 $f_{\text{fair}}(\mathcal{T}; \hat{\theta}) $ is instantiated by the fairness metrics in the validation set $\mathcal{T}$.  Specifically, consider binary sensitive attribute $g \in \{0,1\}$  and the predicted class probabilities $\hat{y}$. The group fairness metrics,  i.e., demographic parity (DP) can be quantified by $f_{\text{DP}}(\mathcal{T}; \hat{\theta})= \big| \mathbb{E}_{\mathcal{T}}[\hat{y} \mid g=0]- \mathbb{E}_{\mathcal{T}}[\hat{y} \mid g=1] \big|$,  while equal opportunity (EOP) can be quantified by $f_{\text{EOP}}(\mathcal{T}; \hat{\theta})  = \big| \mathbb{E}_{\mathcal{T}}[\ell(z ; \theta) \mid g=1, y=1] -\mathbb{E}_{\mathcal{T}}[\ell(z ; \theta) \mid g=0, y=1] \big|$. 
 Similar to the interpretation of utility, a negative value of $\mathcal{I}_{\text {fair}}\left(z_j;-1 \right)$ indicates a lower $f_{\text{fair}}(\mathcal{T}; {\theta}) $ on a model trained without sample $z_j$, implying an improvement in fairness metric.

\textbf{Robustness:} 
 $ \mathcal{I}_{\text{robust}}(z_j;-1 )\!= \! \sum_{ \tilde{\mathcal{T}}} \!  \nabla_{\! {\theta}} \ell(\tilde{z} ; \hat{\theta})^{\!\top} \mathbf{H}_{\hat{\theta}}^{-\!1}  \nabla_{{\theta}} \ell(z_j; \hat{\theta})$.
For a perturbed dataset $\tilde{\mathcal{T}}$ with adversarial sample $\tilde{z} = z - \gamma \frac{\hat{\theta}^{\top} z + b}{\hat{\theta}^{\top}\! \hat{\theta}} \hat{\theta}$ crafted from sample $z \! \in \! \mathcal{T}$, where $\hat{\theta}$ denotes a linear model, $b\! \in \! \mathbb{R}$ is intercept, and $\gamma \!>\! 1$ controls the magnitude of perturbation. Since the decision boundary is a hyperplane, adversaries can change the prediction by adding minimal perturbations to move each sample orthogonally. A negative value of $\mathcal{I}_{\text {robust}}\left(z_j;-1 \right)$ indicates a lower $f_{\text{fair}}(\mathcal{T}; {\theta}) $ on a model trained without sample $z_j$, implying an improvement in robustness metric.

\vspace{-0.6em}
\section{Proposed Approaches}
\vspace{-0.5em}

\label{SEC-4}
We first introduce the weighted influence functions in \cref{SEC-4-1}, analytically deriving the weights by solving a convex quadratic programming problem in \cref{SEC-4-2}. This foundation enables fine-grained model adjustments through a soft-weighted machine unlearning framework, as detailed in \cref{SEC-4-3}. We then highlight its broad applicability and compatibility with diverse unlearning paradigms in \cref{SEC-4-4}.

\vspace{-0.5em}
\subsection{Step 1: Weighted Influence Function}
\label{SEC-4-1}
Due to the challenges of directly removing samples stated in \cref{SEC-1}, we do not  explicitly set the binary weighting factor $\epsilon \!= \! -1$ or $\epsilon \!= \! 0$ as in previous machine unlearning works \cite{chen2024fast,zhang2023recommendation} when perturbing \cref{Eq: EMR}, but instead introduce the following weighted influence function:
\begin{itemize}[leftmargin=0.3343cm, itemindent=0cm]
\vspace{-0.7em}
    \item[$\bullet$] \textbf{Weighted Influence Function on the Utility Metric:}
\vspace{-0.6em}
    \begin{equation}
    \label{Eq: Weighted Utility}
 \!\!\! \mathcal{I}_{\text {util}}\left(z_j;\epsilon_j \right)= - \epsilon_j  \sum_{z \in \mathcal{T}} \nabla_{{\theta}} \ell(z ; \hat{\theta})^{\top} \mathbf{H}_{\hat{\theta}}^{-1} \nabla_{{\theta}} \ell(z_j; \hat{\theta}).
\end{equation}

\vspace{-2em}
    \item[$\bullet$]  \textbf{Weighted Influence Function on the Fairness Metric:}
 \vspace{-0.4em}
\begin{equation}
\label{Eq: Weighted Fairness}
\!\!\!\!\!\! \mathcal{I}_{\text {DP/EOP}}(z_j;\epsilon_j) \!\!=\!\!  - \epsilon_j \! \nabla_{\!{\theta}} f_{\text{DP/EOP}}(\mathcal{T} ; \hat{\theta})^{\! \top} \! \mathbf{H}_{\hat{\theta}}^{-\!1} \nabla_{\!{\theta}} \ell(z_j; \hat{\theta}) .
\end{equation}

 \vspace{-1em}
    \item[$\bullet$] \textbf{Weighted Influence Function on the Robustness Metric:}
    \vspace{-0.5em}
\begin{equation}
\label{Eq: Weighted Robust}
\!\!\!\!\!  \mathcal{I}_{\text {robust}}\left(z_j;\epsilon_j \right)= - \epsilon_j \sum_{ \tilde{z} \in \tilde{\mathcal{T}}}  \nabla_{\! {\theta}} \ell(\tilde{z} ; \hat{\theta})^{\!\top}\mathbf{H}_{\hat{\theta}}^{-\! 1} \nabla_{{\theta}} \ell(z_j; \hat{\theta}) .
\end{equation}
\end{itemize}

\vspace{-1em}
 Note that for each of the above functions, $\mathcal{I}\left(z_j; \epsilon_j \right) = -\epsilon_j \mathcal{I}\left(z_j; -1 \right)$, where $\epsilon_j$ is not binary ($\epsilon \!= \! -1$ or $0$), but can be optimized based on $\mathcal{I}\left(z_j; \! -1 \right)$ obtained by validation set in the next step. 
 

\vspace{-0.4em}
\subsection{Step 2: Weights Discovery via Optimization}
\vspace{-0.4em}
\label{SEC-4-2}
The goal is to discover $\bm{\epsilon}$ that ensure the model's utility is not adversely affected by the unlearning algorithms across different tasks, colloquially, mitigating over-unlearning.
We formulate it as a convex quadratic programming problem:

\begin{subequations}
\label{Eq: optim} 
\begin{align}
\operatorname{minimize}_{\bm{\epsilon}} & \quad \sum_{i=1}^n  \mathcal{I}_{\text{metric}}\left(z_i; \epsilon_i \right) + \lambda \Vert {\bm{\epsilon}} \Vert^{2}_{2}, \label{Eq: obj}\\
    \text{subject to} 
        &  \quad \sum_{i=1}^n  \mathcal{I}_{\text{metric}} \left(z_i; \epsilon_i \right) \geq - \Delta, \label{Eq: const1}
        \\  
        & \quad \sum_{i=1}^n  \mathcal{I}_{\text{util}}\left(z_i; \epsilon_i \right)\leq 0.          
 \label{Eq: const2}
\end{align}
\end{subequations}

\vspace{-1em}
In \cref{Eq: obj}, depending on the target task, the first term $ \mathcal{I}_{\text{metric}}\left(z_i; \epsilon_i \right) $ represents either $ \mathcal{I}_{\text{fair}}\left(z_i; \epsilon_i \right) $ or $ \mathcal{I}_{\text{robust}}\left(z_i; \epsilon_i \right) $. The second term seeks to penalize changes in the weights $ \bm{\epsilon} $, ensuring that perturbations remain infinitesimal. In the first subjective \cref{Eq: const1}, $ \Delta $ quantifies the current model's fairness $ f_{\text{fair}}(\mathcal{T}; \hat{\theta}) $ or robustness $\sum_{ \tilde{z} \in \tilde{\mathcal{T}}}  \nabla_{\! {\theta}} \ell(\tilde{z} ; \hat{\theta})^{\!\top}$. The constraint $-\Delta$ provides lower bound to prevent over-correction, which could lead to reverse bias or vulnerability. The second subjective \cref{Eq: const2} ensures that the resulting weights preserve the model’s utility without compromise. Building on the problem setting, we can either use a linear solver (e.g., Gurobi \cite{gurobi}) or derive the piecewise‐defined analytical solutions for the different active‐set cases to obtain the optimal weight,

\vspace{-1em}
\begin{equation}
\label{Eq: Solutions}
   \bm{\epsilon}^* \!=\! \left\{
\begin{aligned}
&{\bm{\mathcal{I}}_{\text{metric}} }/{(2\lambda)},&\!\!\!\!   \text{Condition 1,}
\\
&{\Delta}/{| \bm{\mathcal{I}}_{\text{metric}} |^2}\cdot \bm{\mathcal{I}}_{\text{metric}},&\!\!\!\! \text{Condition 2,}\\
&
{\big(\bm{\mathcal{I}}_{\text{metric}} \!-\!({\bm{\mathcal{I}}^{\top}_{\text{metric}}\bm{\mathcal{I}}_{\text{util}}})/{|\mathcal{I}_{\text{util}}|^2} \cdot \bm{\mathcal{I}}_{\text{util}}\big)}/{(2\lambda)},&\!\!\!\!   \text{Condition 3,}
\\
& \frac{\Delta\big(\bm{|\mathcal{I}}_{\text{util}}|^2  \bm{\mathcal{I}}_{\text{metric}} -\bm{\mathcal{I}}^{\top}_{\text{metric}}\bm{\mathcal{I}}_{\text{util}}  \bm{\mathcal{I}}_{\text{util}} \big)}{\bm{|\mathcal{I}}_{\text{metric}}|^2|\bm{\mathcal{I}}_{\text{util}}|^2 - (\bm{\mathcal{I}}^{\top}_{\text{metric}}\bm{\mathcal{I}}_{\text{util}})^2 },&\!\!\!\! \text{Condition 4.}
\end{aligned}
\right.
\end{equation}

\vspace{-0.9em}
Where $\bm{\mathcal{I}} \! =\!\left( \mathcal{I}\left(z_1; -1 \right),\! \cdots,\!  \mathcal{I}\left(z_n; -1 \right)\right)^{\top}$ for samples $\{z_i\}_{i=1}^{n}$. See \cref{APPENDIX-AS} for more details. 

\vspace{-0.2em}
\subsection{Step 3: Weighted Model Unlearning}
\vspace{-0.2em}
\label{SEC-4-3}
Given the aforementioned optimization yielding weights $\bm{\epsilon}^*$, the influence function based unlearning algorithm can be updated in the following closed-form expression:
\vspace{-0.5em}
\begin{equation}
\label{Eq: Weighted Update}
    \hat{\theta}(\mathcal{D} ; \bm{\epsilon}^*) -\hat{\theta}(\mathcal{D} ; 0) \approx  -\frac{1}{n} \sum_{i \in \mathcal{D}} \epsilon^*_i \mathbf{H}_{\hat{\theta}}^{-1}     \nabla_{{\theta}} \ell(z_i ; \hat{\theta}).
\end{equation}

\vspace{-1.2em}
For the majority of classification models, \cref{Eq: Weighted Update} can efficiently update the non-convex model's convex surrogate, i.e., by treating the earlier layers as feature extractors and updating the final fully connected linear layer, and its effectiveness has been demonstrated in many studies, such as, \cite{chen2024fast,DBLP:conf/iclr/ChhabraLM024,DBLP:conf/icml/GuoGHM20,DBLP:conf/icml/KohL17}. Nevertheless, for generative models, the strategies outlined in the footnote of \cref{SEC-3} may not be as effective. In practice, for high-dimensional non-convex models, the statistical noise introduced by estimation can degrade the numerical stability of second-order information, diminishing its potential advantages. As a result, a more practical approach to updating the model is to use a diagonal matrix $\sigma \mathbf{I}$ with a constant $\sigma$ to approximate the inverse of Hessian, and scaling it by the gradient variance as  $\hat{\theta}(z_j ;\epsilon^*_j) \!-\!\hat{\theta}(z_j ;0) \approx \!-\!\frac{\epsilon^*_j}{n} \sigma \mathbf{I} \cdot \nabla_{{\theta}} \ell(z_j ; \hat{\theta}).$
The constant ${ \sigma}/{n}$ can be interpreted as learning rate $\eta$ and estimate $\hat{\theta}(z_j ;\epsilon^*_j)$ through multiple update rounds indexed by $t$,
\begin{equation}
\label{Eq: Weighted FT/GA}
    {\theta}_{t+1}(z_j ;\epsilon^*_j)-{\theta}_{t}(z_j ;0) = -\epsilon^*_j \cdot \eta_{t}   \nabla_{{\theta}} \ell(z_j ; \hat{\theta}) .
\end{equation}
As can be seen in \cref{Eq: Weighted FT/GA}, the soft-weighted scheme can be naturally applied to other unlearning algorithms, e.g., fine-tuning and gradient ascent algorithms, which are currently popular cutting-edge methods in both LLM unlearning \cite{DBLP:conf/acl/JangYYCLLS23,DBLP:journals/corr/abs-2310-10683,DBLP:journals/corr/abs-2404-05868} and non-LLM unlearning \cite{DBLP:conf/nips/KurmanjiTHT23}.

\vspace{-0.5em}
\subsection{Soft-Weighted Unlearning Framework}
\vspace{-0.5em}
\label{SEC-4-4}

To further explore the applicability of the soft-weighted scheme, we elaborate on its relationship with previous baseline methods. Specifically, we define the weight of the forgetting sample as $\epsilon_f$ and the weight for the remaining sample as $\epsilon_r$. In this context, the previous hard-weighted fine-tuning algorithm can be viewed as a special case of our scheme where $\epsilon_f=0$ and $\epsilon_r=1$, while the ascent algorithm represents another special case where $\epsilon_f=-1$ and $\epsilon_r=0$. Since each sample contributes differently to the model, assigning uniform weights can result in the loss of crucial information for prediction, highlighting the issue of over-unlearning as discussed in \cref{SEC-1}.  In contrast, the soft scheme aligns with our intuition: mitigating highly detrimental effects while amplifying beneficial ones.
Moreover, we empirically demonstrate that soft-weighted scheme can also be effectively applied to other heuristic unlearning algorithms, such as Fisher \cite{DBLP:conf/cvpr/GolatkarAS20} or  Teacher-Student Formulation \cite{DBLP:conf/nips/KurmanjiTHT23} et al. Please refer to \cref{APPENDIX-Weighted Unlearning} for details of the soft-weighted version of unlearning algorithms.

Accordingly, we propose the \textbf{Soft-Weighted Unlearning Framework} in \cref{Alg: Framework} to effectively address the over-unlearning challenges commonly encountered in existing non-privacy-oriented tasks, such as bias mitigation and robustness enhancement. This framework introduces a finer-gain approach to unlearning by assigning differentiated weights to samples based on their contributions to the model's objective. Specifically, samples that positively contribute to the objective function are given higher weights, while those that conflict with it are assigned lower weights.  The process of model correction is systematically structured into the following three key steps:

\vspace{-0.8em}
\begin{minipage}[t]{0.38\textwidth}
\vspace{0pt}
    \textbf{Step 1: Influence Evaluation.} We use \cref{Eq: Weighted Fairness,Eq: Weighted Robust} to evaluate the fairness or robustness impact of removing each sample on validation set.  In contrast to previous work \cite{chen2024fast} on fairness, we also use \cref{Eq: Weighted Utility} to evaluate utility. \\
\textbf{Step 2: Weights Optimization.} Based on the results from Step 1, we solve the optimization problem in \cref{Eq: optim} to obtain a set of optimal weights for the training dataset. \\
\textbf{Step 3: Model Correction.} A straightforward way to update the model is through \cref{Eq: Weighted Update}. Nevertheless, our framework is not limited to influence-function-based methods; other unlearning algorithms can also leverage the weights obtained in Step 2 to perform model correction.
\end{minipage}
\hfill
\begin{minipage}[t]{0.577\textwidth} 
\begin{algorithm}[H] 
\caption{Soft-Weighted Unlearning Framework}
\label{Alg: Framework}
\KwIn{Model $\hat{\theta}$, Training Dataset $\mathcal{D}$,  Validationa and Testing Dataset $\mathcal{T}$, Adversarial Samples $\tilde{z}  \!\in \! \tilde{\mathcal{T}}$ }
\textbf{\# Step 1: Influence Evaluation.} \\
\For{each sample $z_i \in \mathcal{D}$}{
    Evaluate influence of $z_i$ on validation set;\\
    Utility: $ \mathcal{I}_{\text {util}}\left(z_i;-1 \right)$ $\leftarrow$ \cref{Eq: Weighted Utility}.\\
    Fairness: $ \mathcal{I}_{\text {fair}}\left(z_i;-1 \right)$ $\leftarrow$ \cref{Eq: Weighted Fairness}.\\
    Robustness: $ \mathcal{I}_{\text {robust}}\left(z_i;-1\right)$ $\leftarrow$ \cref{Eq: Weighted Robust}.
}
\textbf{\# Step 2: Weights Optimization.} \\
Weights $ \{\epsilon_i^*\}_{i=1}^n$ $\leftarrow$ \cref{Eq: Solutions}\\
\textbf{\# Step 3: Model Correction.} \\
\If{$f \leftarrow$ $f_{\text{fair}}(\mathcal{T}; {\theta}) $ or $ \sum_{ \tilde{z} \in \tilde{\mathcal{T}}}  \nabla_{ {\theta}} \ell(\tilde{z} ; {\theta})$ $\leq \delta$}{
$\theta$ $\leftarrow$ \cref{Eq: Weighted Update} or Other Unlearning Algorithms
}
\KwOut{$\theta$}
\end{algorithm}
\end{minipage}

\vspace{-0.64em}
\section{Experiments and Discussion}
\vspace{-0.73em}
\label{SEC-5}
In this section, we conduct two types of experiments to evaluate our findings comprehensively. The first explanatory experiments in \cref{SEC-5-1}, designed to validate the rationale behind motivation discussed in \cref{SEC-1} and methodology presented in \cref{SEC-4}.
The second is applied experiments in \cref{SEC-5-2}, which assess the performance of the soft-weighted framework outlined in \cref{Alg: Framework} in addressing specific challenges, including bias mitigation and robustness improvement. We first estimate the influence of each training sample using a validation dataset prior to unlearning, and then compute the utility, robustness and fairness metrics on the test dataset after the unlearning process is completed.

\textbf{Datasets:} In this work, we follow the experiments setup from \cite{DBLP:conf/iclr/ChhabraLM024} to evaluate on standard fairness and robustness datasets. Specifically, we conducted experiments on \textbf{five real-world datasets}, including two tabular datasets \textbf{UCI Adult} \cite{adult_2},
\textbf{Bank} \cite{DBLP:journals/dss/MoroCR14}, one visual human face dataset \textbf{CelebA} \cite{liu2015faceattributes}, one textual dataset \textbf{Jigsaw Toxicity} \cite{DBLP:journals/corr/abs-1810-01869}. These four datasets are widely adopted benchmarks for evaluating fairness and robustness. In addition, we also evaluate robustness on the \textbf{CIFAR-100} dataset \cite{krizhevsky2009learning}. Further details of datasets can be found in \cref{APPENDIX-Datasets}.

\textbf{Baselines:} We follow the machine unlearning repository in \cite{DBLP:conf/nips/KurmanjiTHT23} with the following \textbf{nine unlearning algorithms}: Gradient Ascent \texttt{(\textbf{GA})} combined with a regularizer Fine-Tuning \texttt{(\textbf{FT})} for utility preservation (Following the definitions in \cite{DBLP:journals/corr/abs-2407-06460}, we denote these combinations as 
$\texttt{\textbf{GA}}_{\texttt{\textbf{FT}}}$.), Influence Function \texttt{(\textbf{IF})} \cite{DBLP:conf/icml/KohL17}, Fisher Forgetting \texttt{(\textbf{Fisher})} \cite{DBLP:conf/cvpr/GolatkarAS20} and NTK Forgetting \texttt{(\textbf{NTK})} \cite{DBLP:conf/eccv/GolatkarAS20}, Teacher-Student Formulation \textbf{\texttt{(SCRUB)}} \cite{DBLP:conf/nips/KurmanjiTHT23} and \texttt{(\textbf{Bad-T})} \cite{DBLP:conf/aaai/ChundawatTMK23}, Freezing and Forgetting Last k-layers Followed by Catastrophic Forgetting-k (\texttt{\textbf{CF-k}}) and Exact Unlearning-k (\texttt{\textbf{EU-k}})\cite{DBLP:journals/corr/abs-2201-06640}, along with their \textbf{S}oft-\textbf{W}eighted \texttt{(\textbf{SW-})} versions. Technical details can be found in \cref{APPENDIX-Weighted Unlearning}.
We evaluated aforementioned unlearning methods on tasks involving fairness and robustness, where we defer EOP to the appendix. 

\textbf{Model:} Similar to \cite{DBLP:conf/iclr/ChhabraLM024}, we train a Logistic Regression (LR) and a Neural Network (NN) with two-layer non-linear structure followed by a linear layer, as well as ResNet-18 and ResNet-50 \cite{he2016deep}. During the retraining or unlearning process, similar to \cite{DBLP:conf/nips/FeldmanZ20}, we consider a faster way to compute influence values: the last layer of the NN or ResNet is treated as a convex surrogate for the non-convex model, and only this part of the parameters is updated.

\subsection{Explanatory Experiments}
\label{SEC-5-1}
\vspace{-1em}
\textbf{\ding{182} Correctness of Influence Evaluations.} Whether using the hard- or soft-weighted scheme, it is 
necessary to evaluate the influence of each sample. However, due to high cost of retraining, it is impractical to train leave-one-out models to determine their actual influence. The soft-weighted framework offers \cref{Eq: Weighted Utility,Eq: Weighted Fairness,Eq: Weighted Robust} to approximate the actual influence on the utility, fairness, and
robustness metrics. The first question naturally is to verify its validity, that is,
\vspace{-0.6em}
\begin{wrapfigure}{r}{0.5855\textwidth}
    \centering
    \vspace{-0.6em}
    \includegraphics[width=\linewidth]{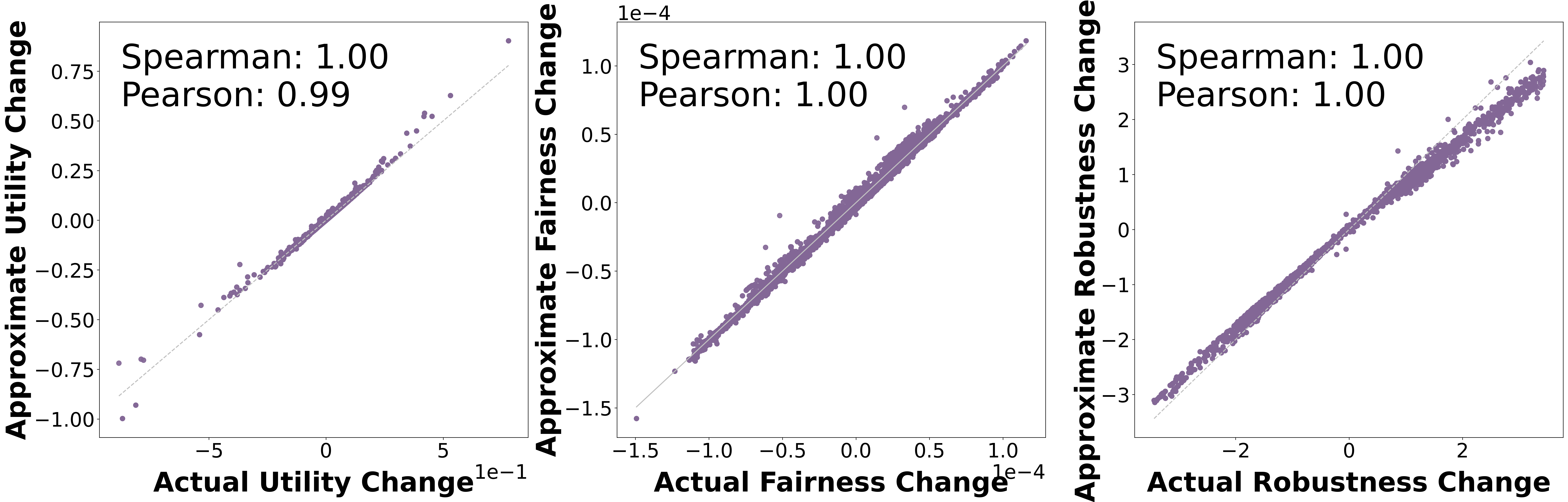}
    \includegraphics[width=\linewidth]{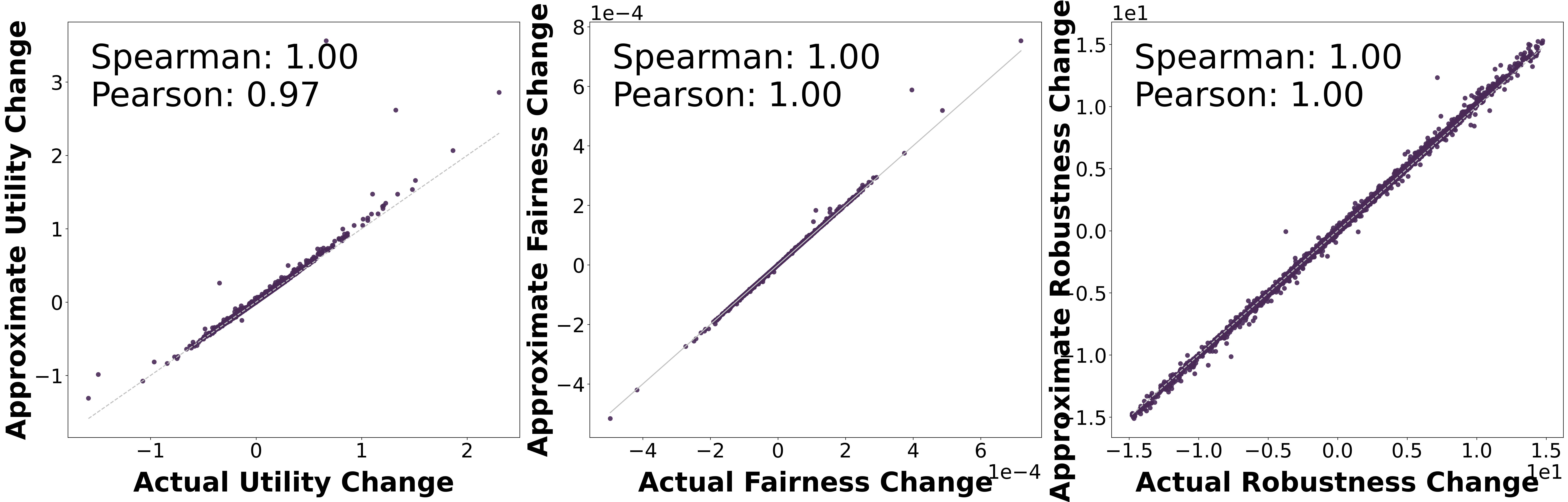}
    \vspace{-1.5em}
    \caption{\textbf{Actual Changes vs. Approximate Changes.} We evaluated the leave-one-out influence for all samples, with the \textbf{First Row} for LR and \textbf{Second Row} for the last layer of NN, on different performance metrics as follows:
    \textbf{(Left)} Model utility (loss on test set), 
    \textbf{(Middle)} fairness (DP loss on test set), 
    \textbf{(Right)} robustness (loss on adversarial sample).}
    \label{Fig: SEC-5.1}
\end{wrapfigure}
\vspace{-0.5em}
\begin{tcolorbox}[before skip=1.5mm, after skip=1.5mm, boxsep=0.0cm, middle=0.0cm, top=0.05cm, bottom=0.05cm,left=0.1cm]
\textit{\textbf{Q}}1: How accurate is the influence evaluation in Step 1?
\end{tcolorbox}

Note that while the validation of influence evaluation has been well established in previous studies, including traditional ML model \cite{DBLP:conf/icml/KohL17,chen2024fast} and non-convex models \cite{DBLP:conf/emnlp/JiaZZLRDK024,zhang2024correctinglargelanguagemodel}, we provide additional justifications for the actual influence and its estimation in Step 1 for the setting in the main text to validate the reliability of influence estimation. 

\textbf{Result.} The influence evaluation values are obtained using \cref{Eq: Weighted Utility,Eq: Weighted Fairness,Eq: Weighted Robust}, while the actual values are obtained by retraining a leave-one-out model for each sample. As illustrated in \cref{Fig: SEC-5.1}, the results from Step 1 exhibit a strong correlation with the actual values in terms of utility, fairness, and robustness metrics, with Spearman \cite{spearman1961proof} and Pearson \cite{Wright1921CorrelationAndCausation} correlation coefficients close to or equal to 1 as depicted in \cref{Fig: SEC-5.1}.

\textbf{\ding{183} Intuition of Over-Unlearning.} After analyzing the counterfactual influence of each sample across different metrics, it becomes essential to understand how adjustments in the weighting strategy influence the model's behavior. In particular, we focus on the intuition behind the transition from the previous hard weights to the softened weights in Step 2. 
\begin{tcolorbox}[before skip=1.5mm, after skip=1.5mm, boxsep=0.0cm, middle=0.0cm, top=0.05cm, bottom=0.05cm,left=0.1cm]
\textit{\textbf{Q}}2: What is the intuition behind using hard weights versus softened weights in Step 2?
\end{tcolorbox}
In \cref{SEC-1}, we discussed three main causes of over-unlearning. To illustrate its advantages, we compare the weighting strategies of hard- and soft-weighted schemes in Step 2.

\begin{figure*}
    \centering{
    {
    \includegraphics[width=0.325\linewidth]{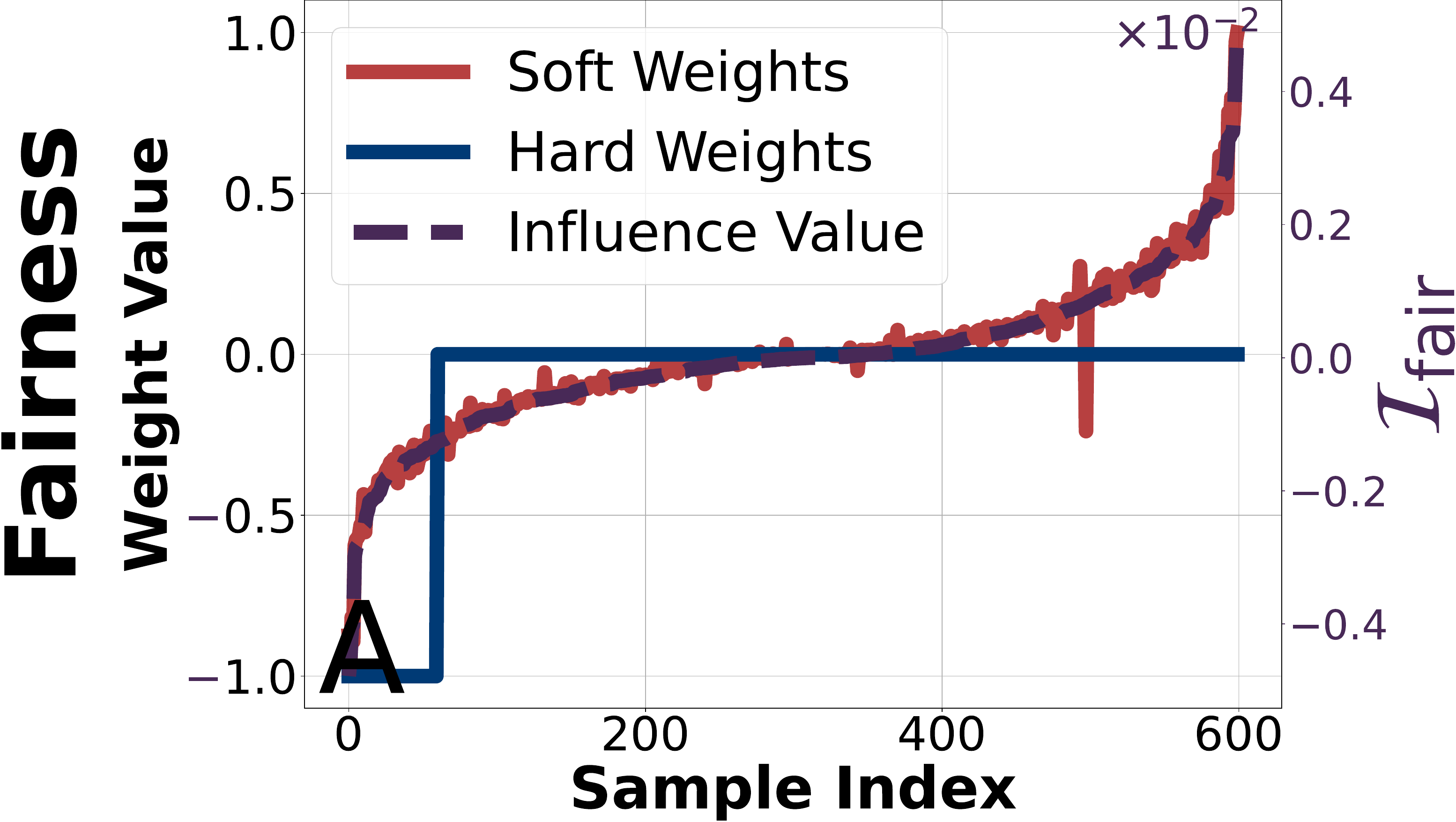}\ \ \ \
    \includegraphics[width=0.285\linewidth]{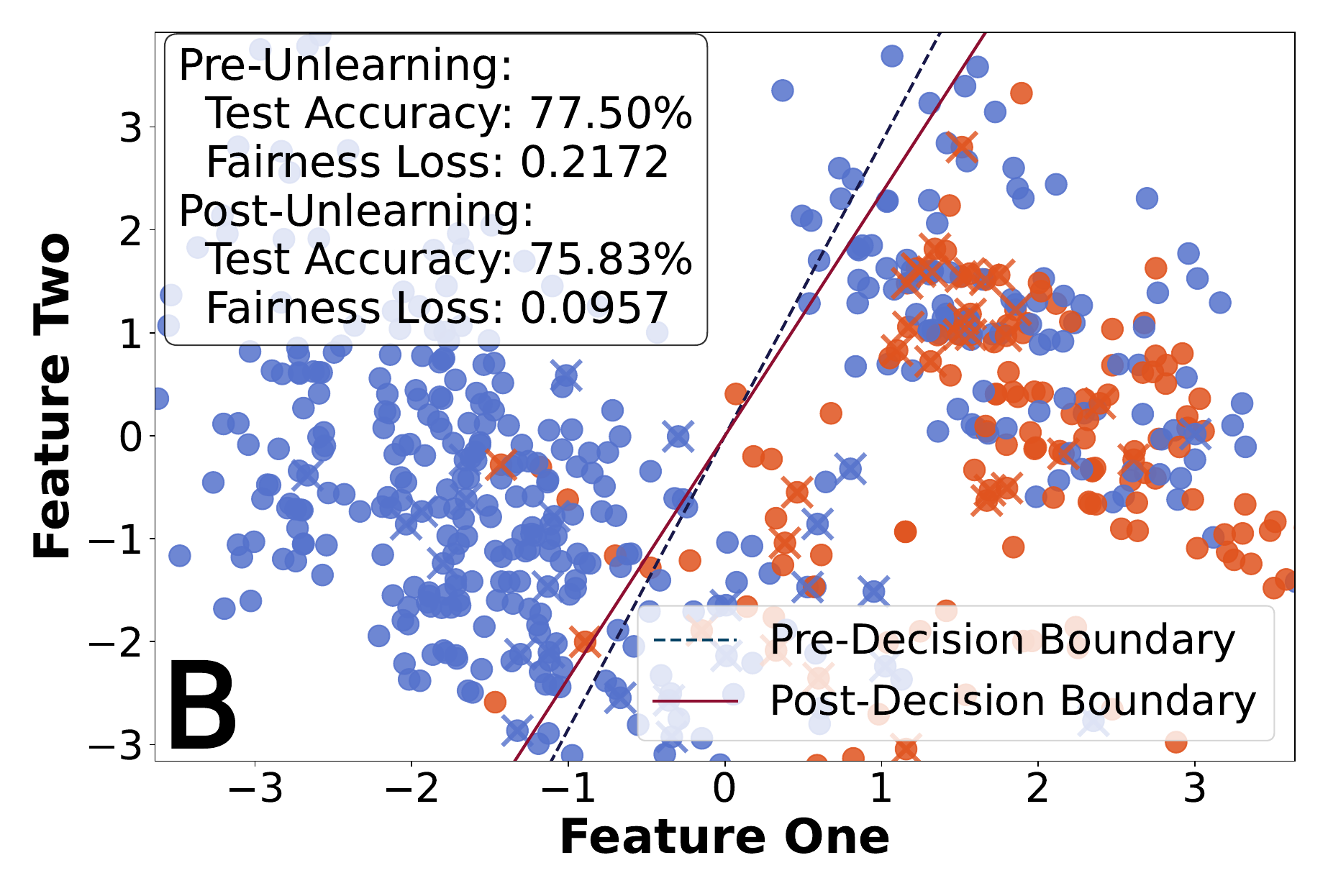}\ \ \
    \includegraphics[width=0.315\linewidth]{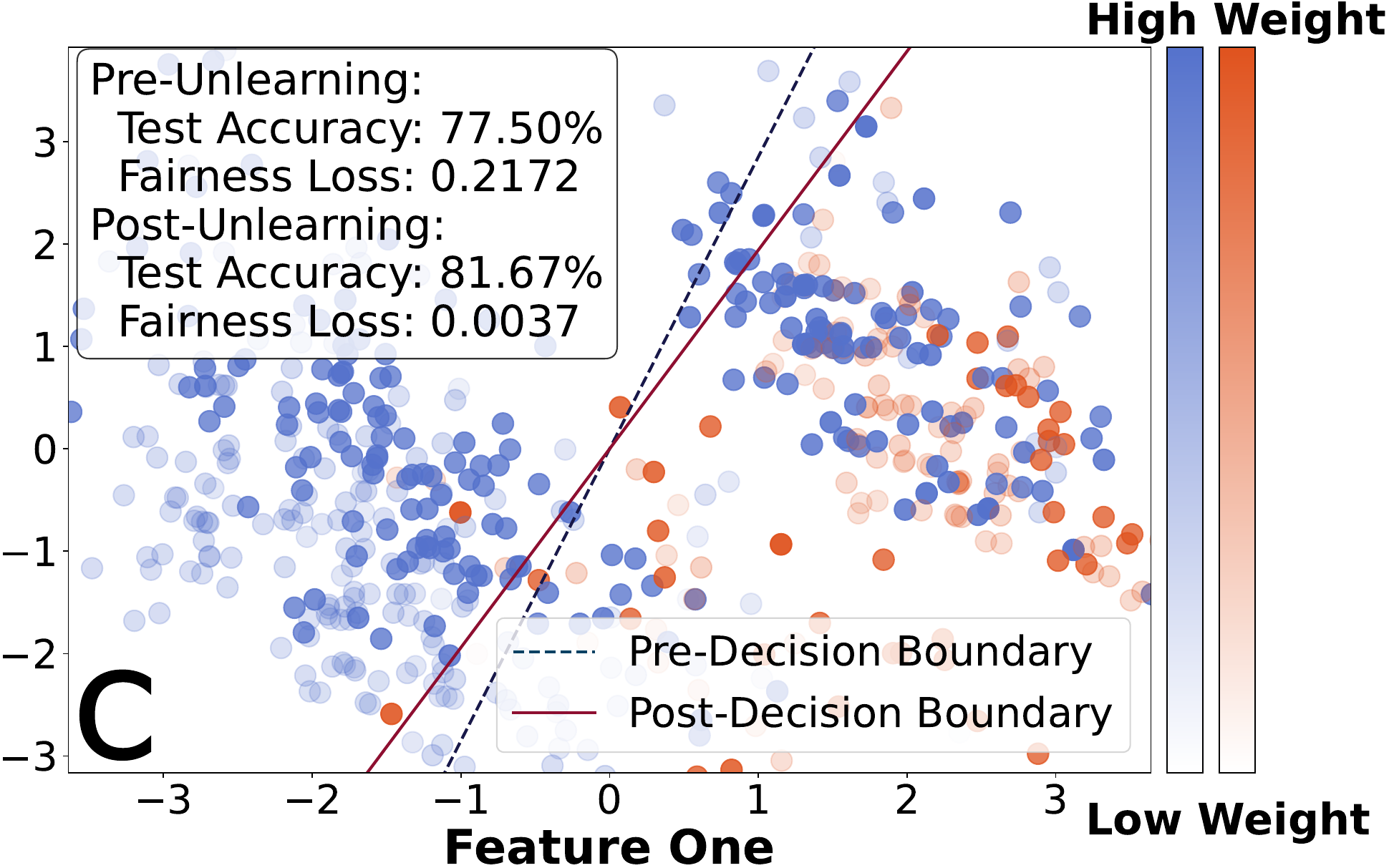} 
    } 
    \\
{ \ \
    \includegraphics[width=0.325\linewidth]{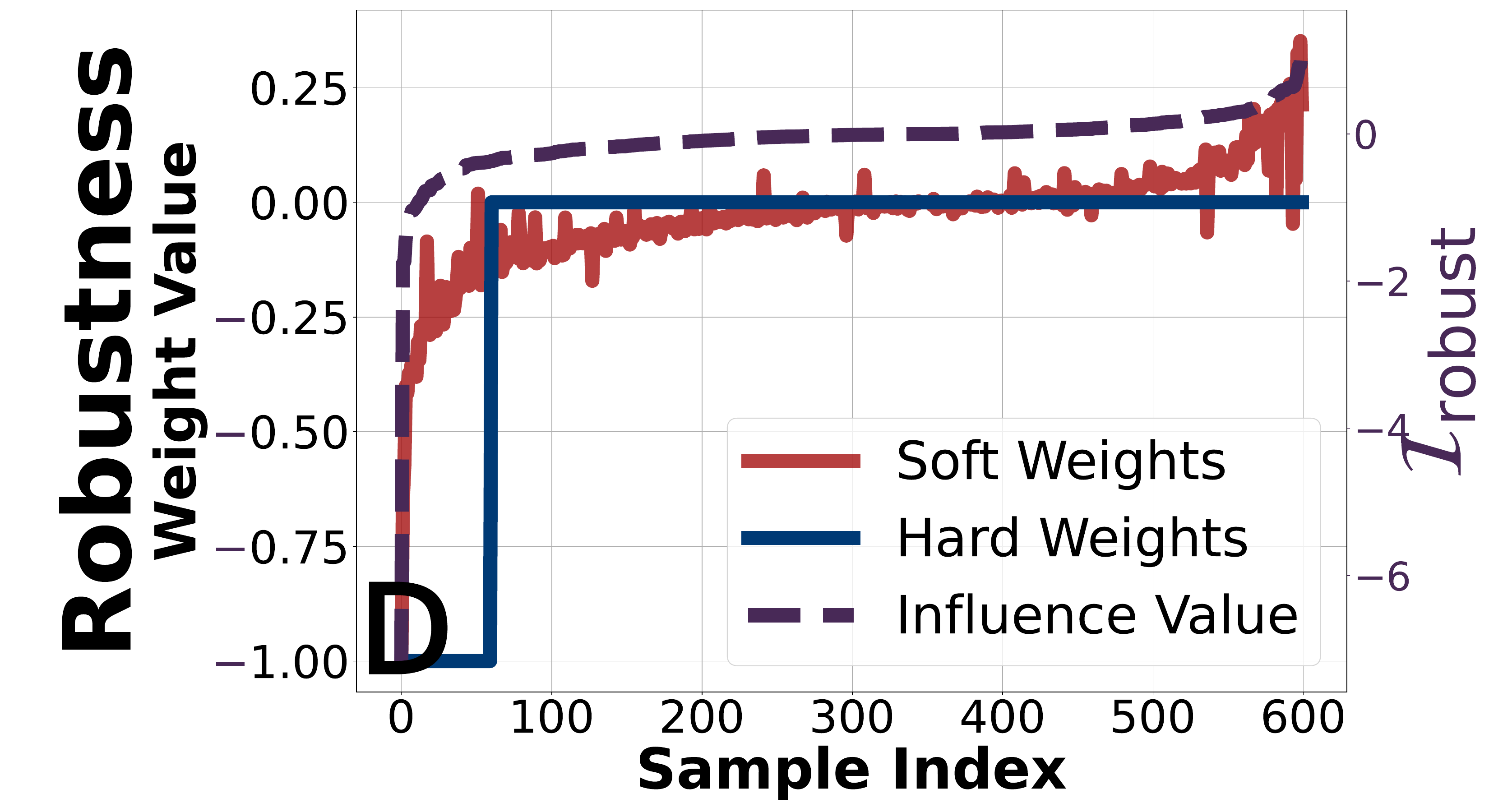}\ \  \ \
    \includegraphics[width=0.285\linewidth]{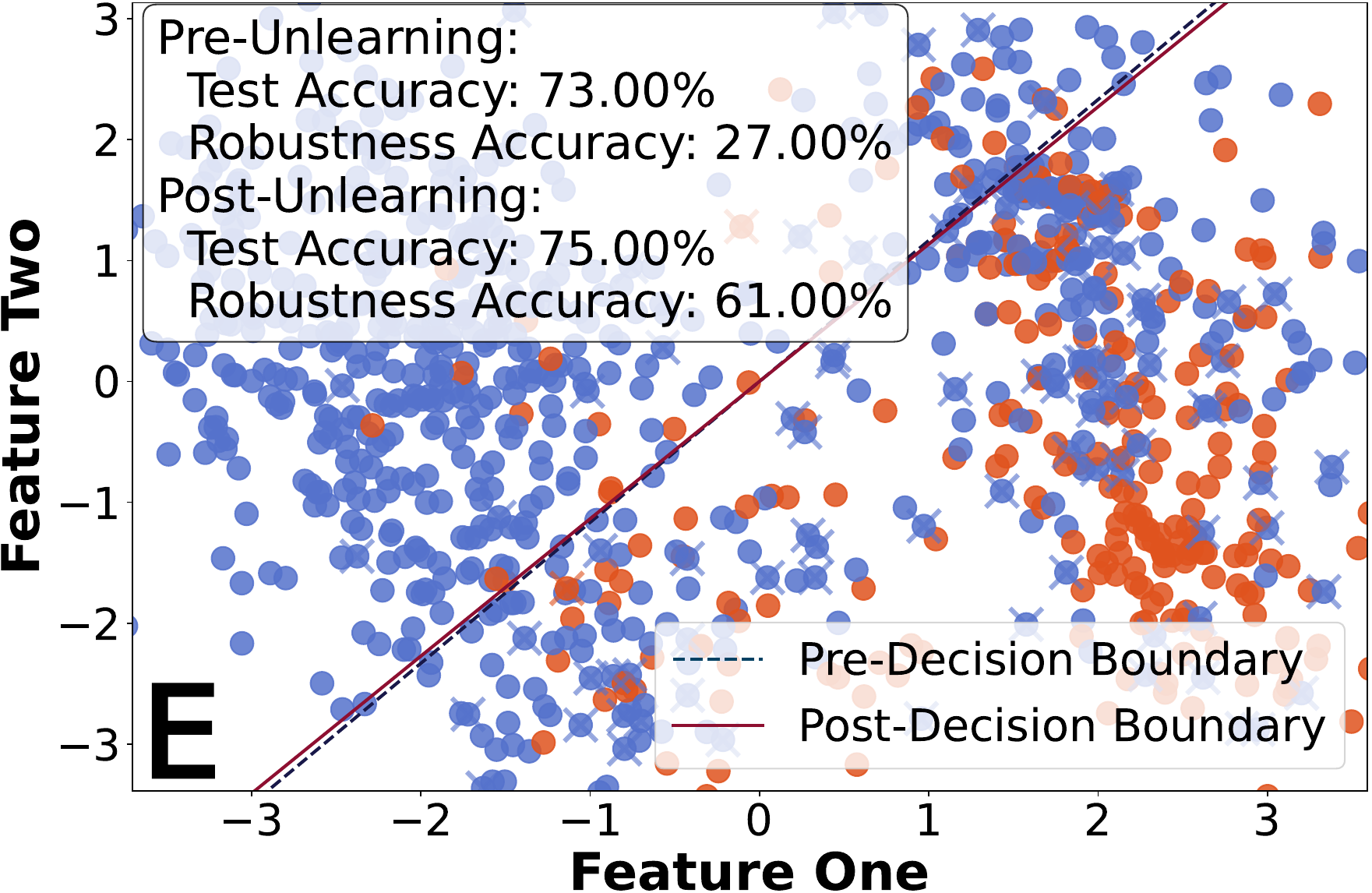}\ \ \
    \includegraphics[width=0.315\linewidth]{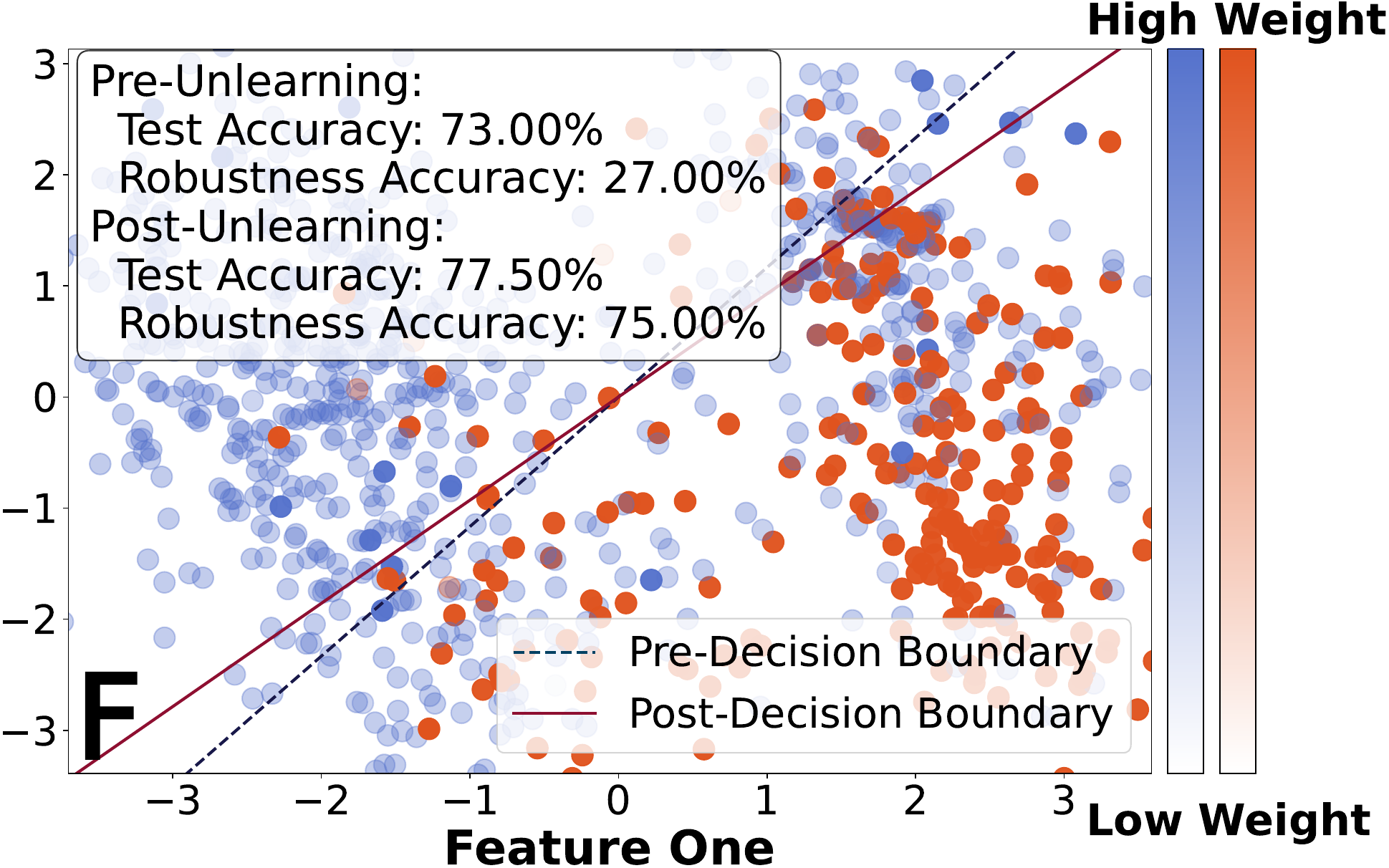} 
    }
}
\caption{\textbf{Hard Weighted Scheme vs. Soft Weighted Scheme.} We use \texttt{\textbf{IF}} as the unlearning method to update model. \textbf{The First Row for Fairness} compares the hard- and soft-weighted schemes: \textbf{A} compares the weighting schemes with corresponding fairness influence values, \textbf{B} presents fairness and utility before and after applying hard-weighted \texttt{\textbf{IF}}, and \textbf{C} shows the same for soft-weighted \texttt{\textbf{IF}}. \textbf{The Second Row for Robustness} follows a similar structure: \textbf{D} compares the weighting schemes and corresponding robustness influence values, \textbf{E} presents robustness and utility before and after applying hard-weighted \texttt{\textbf{IF}}, and \textbf{F} shows the same for soft-weighted \texttt{\textbf{IF}}. Moreover, we use opacity to represent the value of weights.
}
    \label{Fig: SEC-5.3}
\end{figure*}

\textbf{Results.} As shown in \cref{Fig: SEC-5.3} (\textbf{A} and \textbf{D}), hard-weighted schemes  (blue line) involves directly removing the most of biased or adversarially susceptible samples based on their counterfactual influence on fairness $\mathcal{I}_{\text{fair}}$ or robustness $\mathcal{I}_{\text{robust}}$, where the samples are sorted in ascending order based on influence value. Hard weights reflect a clear-cut decision: a sample is either important (e.g., influential) or harmful (e.g., needs to be unlearned).
This is useful when we want to simulate strict removal or control over certain data points, but it neglects both the potential utility of these samples and the residual bias in the remaining data, potentially leading to degraded generalization performance and missed opportunities for further improvements in fairness or robustness.
In contrast, the soft-weighted scheme employs a more refined adjustment mechanism. As illustrated in \cref{Fig: SEC-5.3} (\textbf{A}), the soft weights (red curve) exhibit a smoother distribution compared to the hard weights (blue line). This reflects the scheme's ability to balance the influence of each sample more precisely, ensuring that moderately biased samples are not entirely removed but instead appropriately reweighted. Similarly, \cref{Fig: SEC-5.3} (\textbf{D}) demonstrates how the soft-weighted scheme integrates robustness considerations $\mathcal{I}_{\text{robust}}$, striking a delicate balance between mitigating vulnerabilities and preserving informative samples.

\textbf{\ding{184} Explanation of Model Correction.} 
Building on the insights from Step 2, the next natural step is to explore how soft-weighted adjustments refine the model. A key aspect of this process is to observe the model's decision boundary dynamics. Specifically, we aim to understand:
\begin{tcolorbox}[before skip=1.5mm, after skip=1.5mm, boxsep=0.0cm, middle=0.0cm, top=0.05cm, bottom=0.05cm,left=0.1cm]
\textit{\textbf{Q}}3: How does the decision boundary change before and after soft-weighted correction in Step 3?
\end{tcolorbox}

To better visualize the decision boundary, we use a subset from the training set to obtain a well-trained model.
As shown in \cref{Fig: SEC-5.3} (\textbf{B} and \textbf{E}), the hard-weighted scheme operates with limited information, focusing solely on the most harmful samples while lacking a global view of the other samples. This uniform weighting leads to a lack of information for the remaining data, resulting in limited adjustments.
In contrast, the soft-weighted scheme provides a more holistic understanding of sample importance, allowing the decision boundary to align more closely with higher-weighted samples during classification. Consequently, samples with greater weights are more likely to be correctly classified.
This intuition is clearly reflected in \cref{Fig: SEC-5.3} (\textbf{C} and \textbf{F}): compared to the decision boundary of the original pre-unlearning model, the post-unlearning model’s decision boundary successfully classifies the high-weight samples in the upper-right region while ignoring the low-weight samples in the lower-left region.
This observation aligns well with our intuitions, namely that the unlearning process prioritizes the proper classification of high-weight samples, which are considered more influential in terms of model performance and fairness.

\subsection{Applied Experiments}

\begin{figure*}[t]
    \centering
\includegraphics[width=0.875\linewidth]{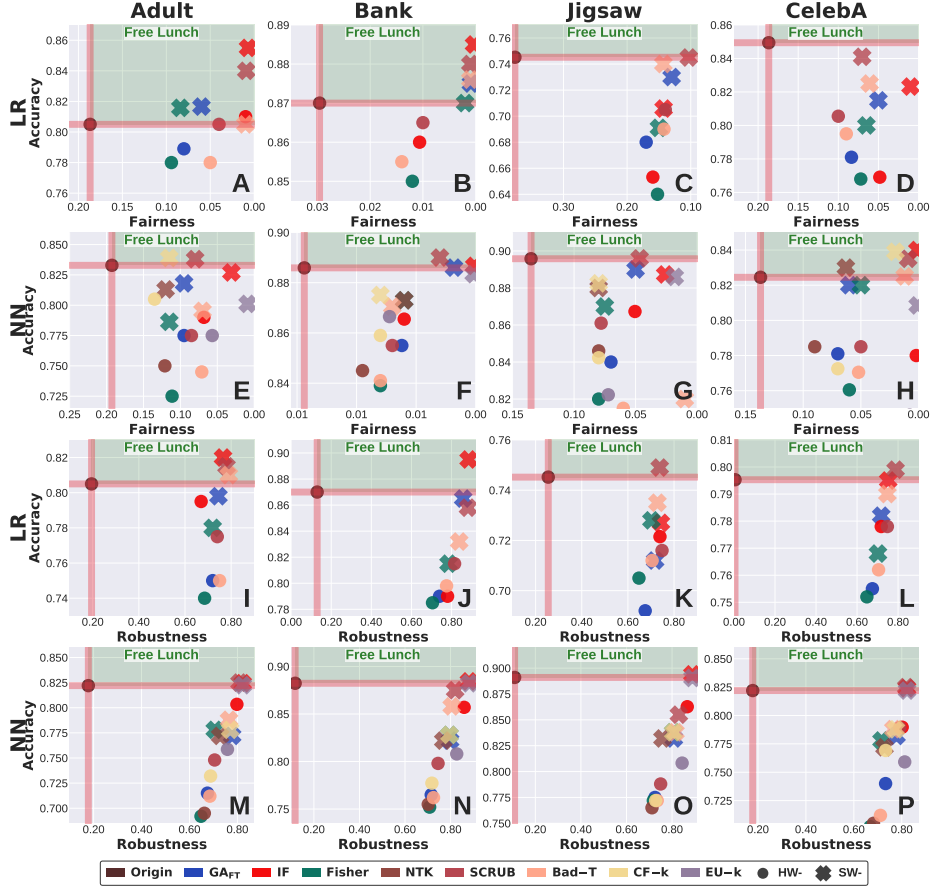}
    \caption{\textbf{Performance on Fairness/Robustness Tasks.} Different colors represent various unlearning algorithms: \ding{108} for the \textbf{H}ard-\textbf{W}eighted scheme and \ding{54} for the \textbf{S}oft-\textbf{W}eighted scheme. \textbf{The First Two Rows} (LR, NN) evaluate utility and fairness metrics, while \textbf{The Last Two Rows} (LR, NN) evaluate utility and robustness metrics across datasets. \textbf{The Green Region} highlights that \textcolor[HTML]{006400}{\textbf{Free Lunch}} cases occurs when unlearning improve both task performance and utility compared to original model. The soft weighting outperforms the hard weighting by enhancing task performance and mitigating decline in utility, even achieving free lunch in some of the unlearning algorithms.}
    \label{Fig: Applied Exeperiments}
\end{figure*}

\label{SEC-5-2}

In this section, we evaluate the performance of different unlearning algorithms under a fixed budget of 30 epochs. For algorithms utilizing gradient descent, we set a learning rate of 0.01, while for those using gradient ascent, we set a learning rate of 0.0005, using full-batch updates. Unless otherwise specified, we use the entire training dataset by default.  For LR, we demonstrate its performance on small datasets using 1,000 training samples from the Adult and Bank datasets. For the hard-weighted scheme, we perform unlearning by iteratively removing the most harmful samples until no further improvement is observed in fairness. It is important to note that unlearning methods' performance may vary across datasets/models depending on hyperparameter choices, and our selected configurations might not be optimal. Our goal is not to assess the superiority of each algorithm, but rather to compare the differences between hard- and soft-weighted schemes, under the same setup and cost constraints. Finally, we evaluate ResNet-50 on CIFAR-100 for robustness and ResNet-18 on CelebA for fairness, with the results deferred to the \cref{APPENDIX-Additional Experiments} due to space limitations.

\textbf{Results.} From \cref{Fig: Applied Exeperiments}, we can observe the following:
(i) In all scenarios \textbf{(A-P)} compared to the hard-weighted method, the soft-weighted scheme outperforms it in terms of target task performance. This improvement stems from optimizing the sample weights through objective \cref{Eq: obj} and constraint in \cref{Eq: const1}. Moreover, considering the constraint in \cref{Eq: const2}, the soft-weighting effectively alleviates utility degradation, which is a limitation often observed in the hard-weighted approach.
(ii) In most scenarios \textbf{(A-B, E-P)} compared to the original model, the soft-weighted scheme not only improves the target task performance but also enhances utility in certain algorithms, which we refer to as the "free lunch" cases in this paper, highlighting the dual improvement in both target performance and utility. (iii) In smaller datasets \textbf{(A-B, I-J)} compared to the original model, the free lunch cases becomes especially pronounced. Intuitively, this is because our method estimates the influence value of each data to compute the weights. In larger datasets, the cumulative estimation error becomes more pronounced, which can lead to a slight utility decline. Finally, compared to hard weighting, soft weighting incurs negligible overhead (<0.03\% runtime increase) to calculate the weights, yet it yields substantial improvements. Due to space constraints, we defer the visualization of runtime results to \cref{APPENDIX-Additional Experiments}.

\section{Conclusion}

We investigate the underlying causes of over-unlearning through counterfactual contribution analysis. To address this challenge, we propose an innovative soft-weighted machine unlearning framework that is simple to apply for non-privacy tasks including but not limited to fairness and robustness.
Specifically, we introduce weighted influence functions, and obtain weights by solving convex quadratic programming problem. In contrast to hard-weighted schemes, the finer-grained soft scheme empirically maintains superior task-specific performance and utility with negligible overhead.

\clearpage

\clearpage

\bibliographystyle{unsrt}
\bibliography{Mybib.bib}


\clearpage

\renewcommand{\contentsname}{Contents of Appendix}
\tableofcontents
\addtocontents{toc}{\protect\setcounter{tocdepth}{3}} 

\appendix

\clearpage
\section*{Limitation and Societal Impacts}
\label{SEC: Limitation and Societal Impacts}

The method presented in this study demonstrates considerable potential across a range of applications, especially within the field of machine unlearning. This research is groundbreaking in its investigation of the underlying causes of over-unlearning in non-privacy tasks, with a specific emphasis on fairness and robustness. By providing insights into these challenges, the study seeks to facilitate the development of more advanced and effective unlearning algorithms.

The proposed framework effectively tackles the problem of over-unlearning, offering support to a diverse array of existing machine unlearning algorithms in navigating their respective challenges. However, it is important to note that while the framework is designed to be broadly applicable, its evaluation is constrained by limited resources and the lack of established benchmarks for assessing fairness and robustness in the context of Large Language Model (LLM) unlearning. Consequently,  the performance of popular LLM unlearning algorithms, for instance, gradient ascent, have not been evaluated within LLMs, leaving the effectiveness of the framework in this domain unverified. Future research should prioritize exploring the applicability and performance of this framework in LLM-related tasks.

Moreover, it is also essential to clarify that this research does not aim to introduce new arguments advocating for algorithmic fairness, as interventions designed to promote fairness do not always align with the intended societal outcomes. This raises ongoing questions about the suitability of concepts like group fairness DP and EOP metrics for evaluating the equity of decision-making systems. An important avenue for future research involves investigating whether the findings of this study can be applied to other fairness concepts, such as individual fairness. Beyond fairness and robustness, the implications of this work extend to critical areas such as the removal of poisoned data and management of outdated data, which warrants further investigation.

\section{Technique Details}
\label{APPENDIX-Technique Details}
We provide a more detailed explanation in \cref{SEC-3} to avoid any misleading interpretations, including an explanation of the influence function and quantitative definitions of fairness and robustness.

\subsection{Influence Function}
The empirical risk minimizer for the training dataset $\mathcal{D}=\{z_i=(\mathbf{x}_i, y_i)\}_{i=1}^{n}$ is given by $\hat{\theta}=\arg \min_{\theta \in \Theta} \frac{1}{n} \sum_{i=1}^n \ell\left(z_i; \theta\right)$. For an empirical risk that is twice-differentiable and strictly convex in the parameter space $\Theta$, we perturb the loss for sample $z_j$ (or alternatively, the training input) by reweighting it with a weight $\epsilon_j \in \mathbb{R}$, as follows:
\begin{equation}
    \hat{\theta}(z_j ;\epsilon_j) = \arg \min _{\theta \in \Theta} \frac{1}{n}  \sum_{i=1}^n \left( \ell\left(z_i; \theta\right) + \epsilon_j \ell\left(z_j ; \theta\right)\right).
\end{equation}
\ding{172} We define the actual change between the empirical risk minimizer trained without sample $z_j$, denoted by $ \hat{\theta}(z_j ;-1)$ and the original empirical risk minimizer, denoted by $ \hat{\theta}(z_j ;0)$ as $\mathcal{I}^*_{\text{param}}(z_j ;-1)=  \hat{\theta}(z_j ;-1) - \hat{\theta}(z_j ;0)$. The influence function, using implicit function theory, can effectively approximate the true change in model parameters.
\begin{equation}
\label{Eq: Appendix-A.1}
\textbf{Parameter Influence:} \quad \mathcal{I}^*_{\text{param}}(z_j ;-1)\approx \mathcal{I}_{\text{param}}(z_j ;-1) \stackrel{\text { def }}{=} - \frac{1}{n}  \left.\frac{d \hat{\theta}(z_j ;\epsilon_j)}{d \epsilon}\right|_{\epsilon=0} =  \frac{1}{n} \mathbf{H}_{\hat{\theta}}^{-1} \nabla_{\theta} \ell(z_j; \hat{\theta}).
\end{equation}
For a function $f$ of interest in the model, such as a utility, fairness and robustness metrics, the actual change in the function $f$ can be expressed as
$
\mathcal{I}^*(z_j; \epsilon) = f(\hat{\theta}(z_j; -1)) - f(\hat{\theta}),
$
where $f(\hat{\theta}(z_j; -1))$ denotes the function value on the retraining empirical risk minimizer, and $f(\hat{\theta})$ denotes the function value on the original empirical risk minimizer. 

\ding{173} For the utility metric, we are interested in the loss on the test dataset $\mathcal{T}$, which is given by
$
\sum_{z \in \mathcal{T}} \ell( z ; \hat{\theta})
$. By applying the chain rule, we can estimate the actual change in the utility metric  of each sample $z_j$,
\begin{equation}
\label{Eq: Appendix-A.2}
\begin{aligned}
    \textbf{Utility Influence:} \quad \mathcal{I}^*_{\text{util}}(z_j ;-1)\approx \mathcal{I}_{\text{util}}(z_j ;-1) & \stackrel{\text { def }}{=} -\left.\frac{d \left(\sum_{z \in \mathcal{T}}  \ell( z ; \hat{\theta})^{\top}\right)
}{d \epsilon}\right|_{\epsilon-0}
    \\
    &= -\frac{d \left(\sum_{z \in \mathcal{T}}  \ell( z ; \hat{\theta})^{\top}\right)
}{d \hat{\theta}(z_j ;\epsilon_j)} \left.\frac{d \hat{\theta}(z_j ;\epsilon_j)}{d \epsilon}\right|_{\epsilon=0} 
    \\
    &=  - \sum_{z \in \mathcal{T}} \nabla_{\theta} \ell( z ; \hat{\theta})^{\top}\left.\frac{d \hat{\theta}(z_j ;\epsilon_j)}{d \epsilon}\right|_{\epsilon=0}
    \\
    & = \sum_{z \in \mathcal{T}} \nabla_{\theta} \ell( z ; \hat{\theta})^{\top} \mathbf{H}_{\hat{\theta}}^{-1} \nabla_{\theta} \ell(z_j; \hat{\theta}).
\end{aligned}
\end{equation}
Therefore, $\mathcal{I}_{\text{util}}(z_j ; -1)$ reflects the change in loss on the test set $\mathcal{T}$. A negative value of $\mathcal{I}_{\text{util}}(z_j ; -1)$ indicates that the retraining empirical risk, obtained without sample $z_j$, results in a lower test set loss compared to the original empirical risk, meaning that the utility improves when sample $z_j$ is removed.

\ding{174} For the fairness metric, we focus on the fairness loss calculated on the test dataset $\mathcal{T}$, which is expressed as $ f_{\text{DP/EOP}}(\mathcal{T} ; \hat{\theta}) $.

As an example, consider a binary sensitive attribute $g \in {0,1}$ and the predicted probabilities $\hat{y}$. \\Demographic Parity (which is also referred to as Statistical Parity) is defined as 
$$ f_{\text{DP}}(\mathcal{T}; {\theta}) = \big| \mathbb{E}_{\mathcal{T}}[\hat{y} \mid g=0] - \mathbb{E}_{\mathcal{T}}[\hat{y} \mid g=1] \big|, $$
and it holds when the likelihood of receiving a positive predicted probabilities $\hat{y}$ (e.g., being classified as a good credit risk) is independent of the sensitive attribute $g \in {0,1}$.  On the other hand, the Equality of Opportunity (EOP) metric is defined by 
$$ f_{\text{EOP}}(\mathcal{T}; {\theta}) = \big| \mathbb{E}_{\mathcal{T}}[\ell(z ; {\theta}) \mid g=1, y=1] - \mathbb{E}_{\mathcal{T}}[\ell(z ; {\theta}) \mid g=0, y=1] \big|, $$
which ensures that the true positive rates are equal across subgroups, thereby offering equal opportunities for all groups. The fairness of the two metrics increases as their absolute values decrease.

Therefore, by applying the chain rule, we can approximate the change in the fairness metric of each sample $z_j$. 
\vspace{-0.1cm}
\begin{equation}
\label{Eq: Appendix-A.3}
\begin{aligned}
    \textbf{Fairness Influence:} \quad \mathcal{I}^*_{\text{fair}}(z_j ;-1)\approx \mathcal{I}_{\text{fair}}(z_j ;-1) & \stackrel{\text { def }}{=} -\left.\frac{d \left(f_{\text{DP/EOP}}(\mathcal{T} ; \hat{\theta}) \right)
}{d \epsilon}\right|_{\epsilon-0}
    \\
    &= -\frac{d \left(f_{\text{DP/EOP}}(\mathcal{T} ; \hat{\theta}) \right)^{\top}
}{d \hat{\theta}(z_j ;\epsilon_j)} \left.\frac{d \hat{\theta}(z_j ;\epsilon_j)}{d \epsilon}\right|_{\epsilon=0} 
    \\
    &=  -  \nabla_{\theta} f_{\text{DP/EOP}}(\mathcal{T} ; \hat{\theta}) ^{\top}
    \left.\frac{d \hat{\theta}(z_j ;\epsilon_j)}{d \epsilon}\right|_{\epsilon=0}
    \\
    & =  \nabla_{\theta} f_{\text{DP/EOP}}(\mathcal{T} ; \hat{\theta})^{\top} \mathbf{H}_{\hat{\theta}}^{-1} \nabla_{\theta} \ell(z_j; \hat{\theta}).
\end{aligned}
\end{equation}
Similarly, $\mathcal{I}_{\text{fair}}(z_j ; -1)$ reflects the change in fairness loss on the test set $\mathcal{T}$. A negative value of $\mathcal{I}_{\text{fair}}(z_j ; -1)$ indicates that the empirical risk after retraining without sample $z_j$, leads to a lower fairness loss than the original empirical risk, which suggests that removing sample $z_j$ improves fairness.

\ding{175} For the robustness metric, we focus on the loss $\sum_{ \tilde{\mathcal{T}}}   \ell(\tilde{z} ; \hat{\theta})^{\top}$ calculated on the perturbed test dataset $\tilde{\mathcal{T}}$ with adversarial sample $\tilde{z} = z - \gamma \frac{\hat{\theta}^{\top} z + b}{\hat{\theta}^{\top}\! \hat{\theta}} \hat{\theta}$ crafted from test sample $z  \in  \mathcal{T}$, where $\hat{\theta}$ denotes a linear model, $b \in  \mathbb{R}$ is intercept, and $\gamma > 1$ controls the magnitude of perturbation. Since the decision boundary is a hyperplane, adversary can change the prediction by adding minimal perturbations to move each sample orthogonally. 

Therefore, by applying the chain rule, we can approximate the change in the robustness metric of each sample $z_j$.
\vspace{-0.1cm}
\begin{equation}
\begin{aligned}
    \textbf{Robustness Influence:} \quad \mathcal{I}^*_{\text{robust}}(z_j ;-1)\approx \mathcal{I}_{\text{robust}}(z_j ;-1) & \stackrel{\text { def }}{=} -\left.\frac{d \left(\sum_{ \tilde{z} \in \tilde{\mathcal{T}}}   \ell(\tilde{z} ; \hat{\theta})\right)
}{d \epsilon}\right|_{\epsilon-0}
    \\
    &= -\frac{d \left(\sum_{\tilde{z} \in \tilde{\mathcal{T}}}   \ell(\tilde{z} ; \hat{\theta})^{\top}\right)
}{d \hat{\theta}(z_j ;\epsilon_j)} \left.\frac{d \hat{\theta}(z_j ;\epsilon_j)}{d \epsilon}\right|_{\epsilon=0} 
    \\
    &=  -\sum_{ \tilde{z} \in \tilde{\mathcal{T}}}    \nabla_{\theta} \ell(\tilde{z} ; \hat{\theta})^{\top}\left.\frac{d \hat{\theta}(z_j ;\epsilon_j)}{d \epsilon}\right|_{\epsilon=0}
    \\
    & = \sum_{ \tilde{z} \in \tilde{\mathcal{T}}}    \nabla_{\theta} \ell(\tilde{z} ; \hat{\theta})^{\top} \mathbf{H}_{\hat{\theta}}^{-1} \nabla_{\theta} \ell(z_j; \hat{\theta}).
\end{aligned}
\end{equation}
Similarly, $\mathcal{I}_{\text{robust}}(z_j ; -1)$ reflects the change in the robustness loss on the perturbed test dataset $\mathcal{T}$. A negative value of $\mathcal{I}_{\text{robust}}(z_j ; -1)$ indicates that the empirical risk after retraining without sample $z_j$, leads to a lower robustness  loss than the original empirical risk, which suggests that removing sample $z_j$ improves robustness.

Correspondingly, when we do not explicitly set $\epsilon=-1$, the weighted influence function is given as follows:
\vspace{-0.1cm}
\begin{itemize}[leftmargin=0.3343cm, itemindent=0cm]
 \item[$\bullet$] \textbf{Weighted Influence Function on Model Parameter:}
 \vspace{-0.1cm}
 \begin{equation}
     \mathcal{I}_{\text {param}}\left(z_j;\epsilon_j \right) = -\frac{1}{n} \sum_{i \in \mathcal{D}} \epsilon^*_i \mathbf{H}_{\hat{\theta}}^{-1}     \nabla_{{\theta}} \ell(z_i ; \hat{\theta})
 \end{equation}
    \item[$\bullet$] \textbf{Weighted Influence Function on Utility Metric:}
    \vspace{-0.1cm}
    \begin{equation}
 \mathcal{I}_{\text {util}}\left(z_j;\epsilon_j \right)= - \epsilon_j  \sum_{z \in \mathcal{T}} \nabla_{{\theta}} \ell(z ; \hat{\theta})^{\top} \mathbf{H}_{\hat{\theta}}^{-1} \nabla_{{\theta}} \ell(z_j; \hat{\theta}) .
\end{equation}
    \item[$\bullet$]  \textbf{Weighted Influence Function on Fairness Metric:}
    \vspace{-0.1cm}
\begin{equation}
 \mathcal{I}_{\text {DP/EOP}}(z_j;\epsilon_j) =  - \epsilon_j  \nabla_{{\theta}} f_{\text{DP/EOP}}(\mathcal{T} ; \hat{\theta})^{ \top}  \mathbf{H}_{\hat{\theta}}^{-1} \nabla_{{\theta}} \ell(z_j; \hat{\theta}) .
\end{equation}
    \item[$\bullet$] \textbf{Weighted Influence Function on Robustness Metric:}
\begin{equation}
 \mathcal{I}_{\text {robust}}\left(z_j;\epsilon_j \right)= - \epsilon_j \sum_{ \tilde{z} \in \tilde{\mathcal{T}}}  \nabla_{ {\theta}} \ell(\tilde{z} ; \hat{\theta})^{\top}\mathbf{H}_{\hat{\theta}}^{- 1} \nabla_{{\theta}} \ell(z_j; \hat{\theta}) .
\end{equation}
\end{itemize}
\subsection{Analytical Solution of Problem \ref{Eq: optim}}
\label{APPENDIX-AS}
The objective function in \cref{Eq: optim} contains the squared $\normltwo$ norm with inequality constraint equation constrain \cref{Eq: const1,Eq: const2}. Let $\bm{\mathcal{I}} =\left( \mathcal{I}\left(z_1; -1 \right), \cdots,  \mathcal{I}\left(z_n; -1 \right)\right)^{\top}$. The problem in \cref{Eq: optim} is equivalent to the following problem:
\begin{subequations}
\begin{align}
    \operatorname{minimize}_{\bm{\epsilon}} & \quad -\bm{\epsilon}^{\top} \bm{\mathcal{I}}_{\text{metric}}  + \lambda \Vert {\bm{\epsilon}} \Vert^{2}_{2}\\
    \text{subject to} 
        &   \quad -\bm{\epsilon}^{\top} \bm{\mathcal{I}}_{\text{util}}  \leq 0 \label{Eq:con1}
        \\
        & \quad \bm{\epsilon}^{\top} \bm{\mathcal{I}}_{\text{metric}}  \leq  \Delta. \label{Eq:con2}
\end{align}
\end{subequations}
We formulate the Lagrangian to obtain the following unconstrained optimization problem:
\begin{equation}
L(\bm{\epsilon}, \beta_1,\beta_2)=
-\bm{\epsilon}^{\top} \bm{\mathcal{I}}_{\text{metric}}  + \lambda \Vert {\bm{\epsilon}} \Vert^{2}_{2} - \beta_1  \bm{\epsilon}^{\top}   \bm{\mathcal{I}}_{\text{util}} + \beta_2 (\bm{\epsilon}^{\top} \cdot \bm{\mathcal{I}}_{\text{metric}} - \Delta),
\end{equation}
where $\beta_1\geq 0$ and $\beta_2\geq 0$ are the dual variables corresponding to \cref{Eq:con1} and \cref{Eq:con2}, respectively.
Note that $\mathcal{I}_{\text{metric}}\left(z_j; \epsilon_j \right) = -\epsilon_j \mathcal{I}_{\text{metric}}\left(z_j; -1 \right)$. The feasible solution $\bm{\epsilon}$ needs to satisfy the following KKT conditions:
\begin{subequations}
\label{Eq: APPENDIX-KKT} 
\begin{align}
  \quad  \nabla_{\bm{\epsilon}} L(\bm{\epsilon},  \beta_1,\beta_2) &  =
 -\bm{\mathcal{I}}_{\text{metric}}  + {2}{\lambda}\bm{\epsilon} - \beta_1   \bm{\mathcal{I}}_{\text{util}} + \beta_2 \bm{\mathcal{I}}_{\text{metric}} =\bm{0}, 
 \\    -\bm{\epsilon}^{\top} \bm{\mathcal{I}}_{\text{util}} &\leq 0,
 \\    \bm{\epsilon}^{\top} \bm{\mathcal{I}}_{\text{metric}}- \Delta &\leq 0, 
 \\  -\beta_1  \bm{\epsilon}^{\top}  \bm{\mathcal{I}}_{\text{util}} &= 0
 \\   \beta_2 (\bm{\epsilon}^{\top} \bm{\mathcal{I}}_{\text{metric}}- \Delta) &= 0
 \\
 \quad \beta_1, \beta_2  &\geq 0
\end{align}
\end{subequations}
We have 
\begin{align}
            \bm{\epsilon}^*
        &= \frac{(1 - \beta_2) \cdot \bm{\mathcal{I}}_{\text{metric}}+\beta_1 \cdot \bm{\mathcal{I}}_{\text{util}}}{2\lambda}.
\end{align}

In the following, we consider four cases based on piecewise‐defined conditions:

\begin{equation*}
\begin{aligned}
        \text{Condition 1:}& \ \ 0 \leq \bm{\mathcal{I}}^{\top}_{\text{metric}}\bm{\mathcal{I}}_{\text{util}} \leq  2\lambda \Delta .
        \\
       \text{Condition 2:}& \ \ 
       |\bm{\mathcal{I}}_{\text{metric}} |^2 \- 2\lambda \Delta  \geq 0,\ \bm{\mathcal{I}}^{\top}_{\text{metric}}\bm{\mathcal{I}}_{\text{util}}  \geq 0.
       \\        
       \text{Condition 3:}&\ \  \bm{\mathcal{I}}^{\top}_{\text{metric}}\bm{\mathcal{I}}_{\text{util}} \leq  0,
       \\
       &(\bm{\mathcal{I}}^{\top}_{\text{metric}}\bm{\mathcal{I}}_{\text{util}} )^2 \geq |\bm{\mathcal{I}}_{\text{util}}|^2 ( | \bm{\mathcal{I}}_{\text{metric}} |^2 - 2\lambda \Delta).
       \\        
       \text{Condition 4:}&\ \  \bm{\mathcal{I}}^{\top}_{\text{metric}}\bm{\mathcal{I}}_{\text{util}} \leq  0,
       \\
       & (\bm{\mathcal{I}}^{\top}_{\text{metric}}\bm{\mathcal{I}}_{\text{util}} )^2 \leq |\bm{\mathcal{I}}_{\text{util}}|^2 ( | \bm{\mathcal{I}}_{\text{metric}} |^2 - 2\lambda \Delta).
       \\        
\end{aligned}
\end{equation*}

\textbf{Case 1:} For $\beta_1=0,\ \beta_2=0 $, we obtain:\\
\textbf{Case 1 condition: }  When $0 \leq \bm{\mathcal{I}}^{\top}_{\text{metric}}\bm{\mathcal{I}}_{\text{util}} \leq  2\lambda \Delta$ and $|\bm{\mathcal{I}}_{\text{metric}}|^2<2\lambda \Delta$, the analytical solution is given as follows:
\begin{equation}
\begin{aligned}
        \bm{\epsilon}^*
        &= \frac{(1 - \beta_2) \cdot \bm{\mathcal{I}}_{\text{metric}}+\beta_1 \cdot \bm{\mathcal{I}}_{\text{util}}}{2\lambda}
        \\&=\frac{\bm{\mathcal{I}}_{\text{metric}}}{2\lambda}. 
\end{aligned}
\end{equation}

\textbf{Case 2:} For $\beta_1=0,\ \beta_2=1- \frac{2 \lambda \Delta}{\bm{|\mathcal{I}}_{\text{metric}}|^2}  \geq 0 $, we obtain:\\
\textbf{Case 2 Condition:} 
$|\bm{\mathcal{I}}_{\text{metric}} |^2 - 2\lambda \Delta  \geq 0,\ \bm{\mathcal{I}}^{\top}_{\text{metric}}\bm{\mathcal{I}}_{\text{util}} \geq  0$, the analytical solution is given as follows:
\begin{equation}
\begin{aligned}
        \bm{\epsilon}^*
        &= \frac{(1 - \beta_2)\bm{\mathcal{I}}_{\text{metric}}+\beta_1\bm{\mathcal{I}}_{\text{util}}}{2\lambda}
        \\
        &=\frac{\Delta}{| \bm{\mathcal{I}}_{\text{metric}} |^2}\cdot \bm{\mathcal{I}}_{\text{metric}}. 
\end{aligned}
\end{equation}

\textbf{Case 3:} For $\beta_1 = -\frac{\bm{\mathcal{I}}^{\top}_{\text{metric}}\bm{\mathcal{I}}_{\text{util}}}{|\mathcal{I}_{\text{util}}|^2} \geq 0,\ \beta_2=0$, we obtain:\\
\textbf{Case 3 Condition:} $\bm{\mathcal{I}}^{\top}_{\text{metric}}\bm{\mathcal{I}}_{\text{util}} \leq  0,\
(\bm{\mathcal{I}}^{\top}_{\text{metric}}\bm{\mathcal{I}}_{\text{util}} )^2 \geq |\bm{\mathcal{I}}_{\text{util}}|^2 ( | \bm{\mathcal{I}}_{\text{metric}} |^2 - 2\lambda \Delta)
$, the analytical solution is:
\begin{equation}
\begin{aligned}
        \bm{\epsilon}^* 
        &=  \frac{(1 - \beta_2)\bm{\mathcal{I}}_{\text{metric}}+\beta_1\bm{\mathcal{I}}_{\text{util}}}{2\lambda}\\
        &= \frac{\bm{\mathcal{I}}_{\text{metric}} -\frac{\bm{\mathcal{I}}^{\top}_{\text{metric}}\bm{\mathcal{I}}_{\text{util}}}{|\mathcal{I}_{\text{util}}|^2} \cdot \bm{\mathcal{I}}_{\text{util}}}{2\lambda}.
\end{aligned}
\end{equation}

\textbf{Case 4:} For $\beta_1 = - \frac{2 \lambda \Delta  \bm{\mathcal{I}}_{\text{metric}}^{\top} \bm{\mathcal{I}}_{\text{util}}}{\big(\bm{|\mathcal{I}}_{\text{metric}}|^2|\bm{\mathcal{I}}_{\text{util}}|^2- (\bm{\mathcal{I}}^{\top}_{\text{metric}}\bm{\mathcal{I}}_{\text{util}})^2  \big) } \geq 0,\ 
\beta_2 = 1- \frac{2 \lambda \Delta |\bm{\mathcal{I}}_{\text{util}}|^2}{\bm{|\mathcal{I}}_{\text{metric}}|^2|\bm{\mathcal{I}}_{\text{util}}|^2- (\bm{\mathcal{I}}^{\top}_{\text{metric}}\bm{\mathcal{I}}_{\text{util}})^2}  \geq 0 $, we obtain:\\
\textbf{Case 4 Condition:} $\bm{\mathcal{I}}^{\top}_{\text{metric}}\bm{\mathcal{I}}_{\text{util}} \leq  0,\ (\bm{\mathcal{I}}^{\top}_{\text{metric}}\bm{\mathcal{I}}_{\text{util}} )^2 \leq |\bm{\mathcal{I}}_{\text{util}}|^2 ( | \bm{\mathcal{I}}_{\text{metric}} |^2 - 2\lambda \Delta)$, the analytical solution is: 
\begin{equation}
\begin{aligned}
        \bm{\epsilon}^* 
        &=  \frac{(1 - \beta_2) \cdot \bm{\mathcal{I}}_{\text{metric}}+\beta_1 \cdot \bm{\mathcal{I}}_{\text{util}}}{2\lambda}\\
        &=\frac{\Delta\Big(\bm{|\mathcal{I}}_{\text{util}}|^2 \cdot \bm{\mathcal{I}}_{\text{metric}}-\bm{\mathcal{I}}^{\top}_{\text{metric}}\bm{\mathcal{I}}_{\text{util}} \cdot \bm{\mathcal{I}}_{\text{util}} \Big)}   {\bm{|\mathcal{I}}_{\text{metric}}|^2|\bm{\mathcal{I}}_{\text{util}}|^2- (\bm{\mathcal{I}}^{\top}_{\text{metric}}\bm{\mathcal{I}}_{\text{util}})^2 }.
\end{aligned}
\end{equation}

\subsection{Weighted Machine Unlearning Algorithms}
\label{APPENDIX-Weighted Unlearning}
In this paper, we follow the experimental repository in \cite{DBLP:conf/nips/KurmanjiTHT23} with the following \textbf{nine unlearning algorithms}: Gradient Ascent \texttt{\textbf{(GA)}}, Fine-Tuning \texttt{\textbf{(FT)}}, Influence Function \texttt{\textbf{(IF)}} \cite{DBLP:conf/icml/KohL17}, Fisher Forgetting \texttt{\textbf{(Fisher)}} \cite{DBLP:conf/cvpr/GolatkarAS20} and NTK Forgetting \texttt{\textbf{(NTK)}} \cite{DBLP:conf/eccv/GolatkarAS20}, Teacher-Student Formulation \texttt{\textbf{(SCRUB)}} \cite{DBLP:conf/nips/KurmanjiTHT23} and \texttt{\textbf{(Bad-T)}} \cite{DBLP:conf/aaai/ChundawatTMK23}, Catastrophic Forgetting-k \texttt{\textbf{(CF-k)}} and Exact Unlearning-k \texttt{\textbf{(EU-k)}} \cite{DBLP:journals/corr/abs-2201-06640}, along with their \textbf{S}oft-\textbf{W}eighted \texttt{\textbf{(SW-)}} versions. Specifically,
for training sample $z_j \in \mathcal{D}$, we define $\mathbf{\epsilon}_r$ as the weight of the remaining data $z_r \in \mathcal{D}_r$ and $\mathbf{\epsilon}_f$ as the weight of the forgetting data $z_f \in \mathcal{D}_f$. The following are the technical details of the different machine unlearning methods:

\ding{172} Gradient Update Methods: $\texttt{\textbf{GA}}$ and  $\texttt{\textbf{FT}}$.\\
$\texttt{\textbf{GA}}$ updates the model by adjusting the parameters according to the negative of the update direction computed from the forgetting dataset, thereby maximizing the loss on the forgetting data $z_f$,
\begin{equation}
    {\theta}_{t+1}(z_f ;-1) = {\theta}_{t}(z_f ;-1) + \eta_{t}   \nabla_{{\theta}} \ell(z_f ; {\theta}_{t}(z_f ;-1)),
 \end{equation}
$\texttt{\textbf{FT}}$ updates the model by adjusting the parameters based on the gradient of the loss function computed over the remaining dataset, optimizing the model to retain knowledge while minimizing the loss on the remaining data $z_r$.
\begin{equation}
    {\theta}_{t+1}(z_r ;-1) = {\theta}_{t}(z_r ;-1) - \eta_{t}   \nabla_{{\theta}} \ell(z_r ; {\theta}_{t}(z_r ;-1)),
 \end{equation}
Therefore, the soft-weighted $\texttt{\textbf{GA}}_{\texttt{\textbf{FT}}}$ can be updated in a manner analogous to weighted gradients update.
\begin{equation}
\label{Eq: Appendix-A.4}
    {\theta}_{t+1}(z_j ;\epsilon_j) = {\theta}_{t}(z_j ;\epsilon_j) + \epsilon_j\cdot \eta_{t}   \nabla_{{\theta}} \ell(z_j ; {\theta}_{t}(z_j ;\epsilon_j)),
 \end{equation}
\ding{173} Closed-form Update Mehtods: \texttt{\textbf{IF}}, \texttt{\textbf{Fisher}},  and 
 \texttt{\textbf{NTK}}. \\
 \texttt{\textbf{IF}} performs a closed-form Newton step to estimate the empirical risk minimizer trained without forgetting data $z_f$.
\begin{equation}
\hat{\theta}(z_f ;-1)-\hat{\theta}(z_f ;0) \approx  \frac{1}{n} \mathbf{H}_{\hat{\theta}}^{-1} \nabla_{{\theta}} \ell(z_f ; \hat{\theta}),
\end{equation}
The \texttt{\textbf{Fisher}} and \texttt{\textbf{NTK}} both require Hessian approximation. \texttt{\textbf{Fisher}} approximates the Hessian using the Fisher Information Matrix. \texttt{\textbf{NTK}} provides a neural tangent kernel (NTK)-based approximation of the training process and uses it to estimate the updated network parameters after forgetting. Formally, \texttt{\textbf{NTK}}, \texttt{\textbf{Fisher}}, and \texttt{\textbf{IF}} are similar and can be interchangeable in special cases. For instance, in the case of an $\normltwo$ loss, the NTK model \texttt{\textbf{NTK}} coincides with the Fisher Matrix.

Therefore, \texttt{\textbf{IF}}, \texttt{\textbf{NTK}}, and \texttt{\textbf{Fisher}} can all be weighted in a manner analogous to the following soft-weighted \texttt{\textbf{IF}},
\begin{equation}
\hat{\theta}(z_f ;\epsilon_f)-\hat{\theta}(z_f ;0) \approx  - \epsilon_f \cdot\frac{1}{n}  \mathbf{H}_{\hat{\theta}}^{-1} \nabla_{{\theta}} \ell(z_f ; \hat{\theta}),
\end{equation}

\ding{174} Teacher-Student (T-S) Framework  Methods: $\texttt{\textbf{SCRUB}}$ and  $\texttt{\textbf{Bad-T}}$.\\
\texttt{\textbf{SCRUB}} considers two sets of teachers: the original model as the "teacher" and the student model. The student is encouraged to stay close to the teacher on the remaining dataset and move away from it on the forgetting dataset. 
\texttt{\textbf{SCRUB}} aims to optimize the following objective function:
\begin{equation}
\label{Eq: Appendix-A.5}
\min _{{\theta}} \frac{\alpha}{|\mathcal{D}_r|} \sum_{z_r \in \mathcal{D}_r} d\left(z_r ; {\theta}(z_f ;-1)\right)+\frac{\gamma}{|\mathcal{D}_r|} \sum_{z_r \in \mathcal{D}_r} f\left(z_r ; {\theta}(z_f ;-1)\right)-\frac{1}{|\mathcal{D}_f|} \sum_{z_f \in \mathcal{D}_f} d\left(z_f ; {\theta}(z_f ;-1)\right).
\end{equation}
where $d\left(z ; {\theta}(z_f ;-1)\right)=D_{\mathrm{KL}}\left(p\left(f\left(z ; {\theta}(z_f ;0)\right)\right) \| p\left(f\left(z ; {\theta}(z_f ;-1)\right)\right)\right)$ is the KL-divergence between the student and teacher output distributions (softmax probabilities) for the sample $z_j$, with hyperparameters $\alpha$ and $\gamma$. Specifically, in \cref{Eq: Appendix-A.5}, the third term involves maximizing the distance between the student and teacher on the forget dataset $\mathcal{D}_f$. The first term is analogous to the third but encourages the student to remain proximal to the teacher on remaining dataset $\mathcal{D}_r$.
Finally, the second term optimizes for the loss on the remaining dataset $\mathcal{D}_r$,

The optimization process alternates between the remaining dataset (\textit{the min-step}) and forgetting dataset (\textit{the max-step}),
\begin{equation}
\textit{\textit{the min-step}:}\ {\theta}(z_r ;-1) \leftarrow {\theta}(z_r ;-1)+\eta \nabla_{{\theta}} \frac{1}{|\mathcal{D}_r|} \sum_{z_r \in \mathcal{D}_r} d\left(z_r ; {\theta}(z_r ;-1)\right).
\end{equation}
\begin{equation}
\textit{\textit{the max-step}:}\ {\theta}(z_f ;-1) \leftarrow {\theta}(z_f ;-1)+\eta \nabla_{{\theta}(z_f ;-1)} \frac{1}{|b|} \sum_{z_f \in b} d\left(z_f ; {\theta}(z_f ;-1)\right) + \gamma f\left(x_r ; {\theta}(z_f ;-1)\right. .
\end{equation}
Considering soft-weighted $\texttt{\textbf{SCRUB}}$, the objective function in \cref{Eq: Appendix-A.5} takes the following form:
\begin{equation}
\min _{{\theta}} \frac{\alpha}{|\mathcal{D}_r|} \sum_{z_r \in \mathcal{D}_r} \epsilon_r \cdot d\left(z_r ; {\theta}(z_f ;\epsilon)\right)+\frac{\gamma}{|\mathcal{D}_r|} \sum_{z_r \in \mathcal{D}_r} \epsilon_r \cdot f\left(z_r ; {\theta}(z_f ;\epsilon_r )\right) + \frac{1}{|\mathcal{D}_f|} \sum_{z_f \in \mathcal{D}_f} \epsilon_f \cdot d\left(z_f ; {\theta}(z_f ;\epsilon_f)\right),
\end{equation}
with following weighted optimization process:
\begin{equation}
\textit{\textit{the min-step}:}\ {\theta}(z_r ;\epsilon_r) \leftarrow {\theta}(z_r ;\epsilon_r)+ \epsilon_r \eta \nabla_{{\theta}} \frac{1}{|\mathcal{D}_r|} \sum_{z_r \in \mathcal{D}_r} d\left(z_r ; {\theta}(z_r ;\epsilon_r)\right).
\end{equation}

\begin{equation}
\textit{\textit{the max-step}:}\ {\theta}(z_f ;\epsilon_f) \leftarrow {\theta}(z_f ;\epsilon_f)+\eta \nabla_{{\theta}(z_f ;\epsilon_f)} \frac{1}{|b|} \sum_{z_f \in b} d\left(z_f ; {\theta}(z_f ;\epsilon_f)\right) + \gamma f\left(x_r ; {\theta}(z_f ;\epsilon_f)\right. .
\end{equation}
\texttt{\textbf{Bad-T}} considers two sets of teachers: the original model as the good teacher and random models as the bad teacher. The student is encouraged to follow the good teacher on the remaining dataset and the bad teacher on the forgetting dataset. 
\begin{equation}
\min_{\theta} 
(1 - y_f)
 * \mathcal{K} \mathcal{L}\left(T_s(x) \| S(x)\right) + y_f *\left(\mathcal{K} \mathcal{L}\left(T_d(x) \| S(x)\right)\right),
\end{equation}
where $T_s(x)$ represents the competent/smart teacher, and $T_d(x)$ is the incompetent/dumb teacher, with $y_f$ being the label of forgetting dataset and $x$ the sample.
The optimization process also alternates between the remaining and forgetting datasets. Due to the similar form of \texttt{\textbf{Bad-T}} and $\texttt{\textbf{SCRUB}}$, we omit the formulation for soft-weighted \texttt{\textbf{Bad-T}}.


\ding{175} Freezing the layers of the neural network Methods: \texttt{\textbf{CF-k}} and  \texttt{\textbf{EU-k}}. 
The \texttt{\textbf{CF-k}} (Catastrophic Forgetting-k) and \texttt{\textbf{EU-k}} (Exact Unlearning-k) methodologies are specifically designed for neural network applications. These approaches operate by first freezing a predefined number of initial layers in the neural architecture, then subsequently either: Fine-tuning the final k layers using the remaining dataset (\texttt{\textbf{CF-k}}), or
Performing complete retraining of the final k layers with the remaining dataset (\texttt{\textbf{EU-k}}).
For implementation convenience, we constrain parameter updates exclusively to the final layer. Consequently, the soft-weighted  \texttt{\textbf{CF-k}} and \texttt{\textbf{EU-k}} adopt the same mathematical formulation presented in \cref{Eq: Appendix-A.4}.

We observe that the overwhelming majority of unlearning algorithms (with the exception of closed-form update methodologies) are predominantly grounded in gradient ascent (GA) and fine-tuning (FT) mechanisms. This analysis delineates their operational specifics through three principal implementation paradigms under fixed epoch constraints:
\begin{itemize}
    \item $\texttt{\textbf{GA}}_{\texttt{\textbf{FT}}}$ employs a two-phase approach, first applying \texttt{\textbf{GA}} on the forgetting dataset for half the total epochs, then \texttt{\textbf{FT}} on the remaining dataset for the latter half. 
    \item \texttt{\textbf{SCRUB}} and \texttt{\textbf{Bad-T}} implement an alternating optimization strategy, interleaving gradient ascent and descent steps using their respective objective functions throughout the training process.
    \item  \texttt{\textbf{CF-k}} conducts \texttt{\textbf{FT}} on remaining dataset across all epochs, contrasting with \texttt{\textbf{EU-k}}'s complete model reinitialization and retraining model.
\end{itemize}


\section{Experiment Details}
\label{Appendix: Experiments}
\subsection{Hardware, Software and Source Code}
The experiments were conducted on an NVIDIA GeForce RTX 4090. The code was implemented in PyTorch 2.0.0 and utilizes the CUDA Toolkit version 11.8. Tests were performed on an AMD EPYC 7763 CPU @1.50GHz with 64 cores, running Ubuntu 20.04.6 LTS.
\newpage

\subsection{Datasets}
\label{APPENDIX-Datasets}
\textbf{Adult Dataset}:
Income prediction dataset with 45,222 samples. Divided into 30,162 training, 7,530 validation, and 7,530 test samples. Gender (male/female) serves as the sensitive attribute for fairness evaluation.

\textbf{Bank Dataset}:
Bank client subscription analysis dataset containing 30,488 entries. Training set (18,292), validation/test sets (6,098 each). Gender (male/female) is designated as the sensitive attribute.

\textbf{CelebA Dataset}:
Facial image dataset comprising 104,163 samples, split into 62,497 training, validation/test sets (20,833  each). Gender (male/female) serves as the sensitive attribute for fairness evaluation.

\textbf{Jigsaw Toxicity Dataset}:
Toxic comment detection corpus with 30,000 social media texts. Training data (18,000), validation/test sets (6,000 each). Ethnicity (Black/Other) serves as the sensitive attribute for fairness evaluation.

\textbf{CIFAR‑100 Dataset:} The CIFAR‑100 dataset is a widely used benchmark for image‑classification research, containing small color images of common objects. All images are 32×32 pixels in RGB format. There are 100 classes, each containing 600 images, grouped into 20 superclasses. The dataset is split into five training sets of 10,000 images each (50,000 total) and one test set of 10,000 images.

\subsection{Additional Experiments}
\label{APPENDIX-Additional Experiments}
This section presents additional experiments, including:
\ding{172} the time distribution of each step in the soft-weighted machine unlearning framework; \ding{173} using the hard-weighted \texttt{(\textbf{IF})} scheme to illustrate how the deletion rate can be selected;
\ding{174} the actual changes in the Equal Opportunity (EOP) fairness metric, the estimated influence values, and the performance of different unlearning algorithms with respect to EOP;
\ding{175} fairness and robustness evaluations on larger models and datasets, specifically ResNet-18 on CelebA for fairness, and ResNet-50 on CIFAR-10 for robustness. \ding{176} Similar to \cref{SEC-1}, we present visualizations of the correlations between fairness/robustness and utility.

\begin{wrapfigure}{r}{0.4\textwidth}
    \centering
    \includegraphics[width=\linewidth]{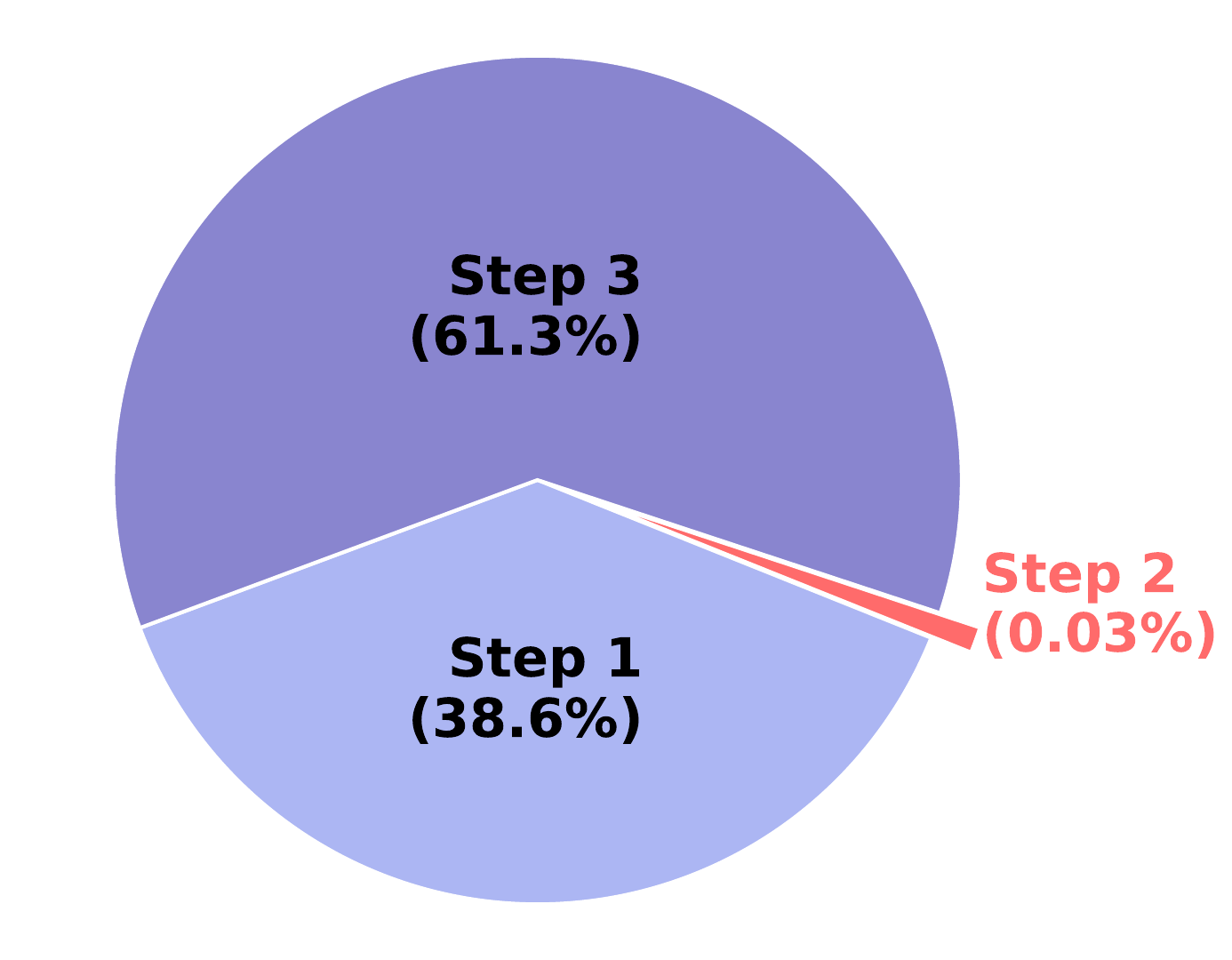}
    \caption{Time cost in each step.  We use \texttt{\textbf{IF}} as the unlearning method to update the model. Step 1 (evaluation) and Step 3 (removal) are common to both hard and soft weighting. Therefore, soft weighting requires only minimal additional time in Step 2.}
    \label{Fig: appendix-time-cost}
\end{wrapfigure}
\subsubsection{Computational Time}
\textbf{First}, \cref{Fig: appendix-time-cost} shows that the time overhead for weight acquisition accounts for only 0.03\% of the total \texttt{\textbf{IF}} unlearning procedure in Step 2. It is noteworthy that the hard weighting framework also necessitates executing Step 1 for sample influence estimation to identify the forgetting dataset, as well as Step 3 to implement the unlearning algorithm. In contrast, the soft weighted machine unlearning framework incurs a smaller overhead in Step 2 to obtain a set of optimal weights while achieving superior performance in Step 3. This underscores the scalability of the soft weighted machine unlearning framework and highlights its strong potential for real-world deployment scenarios.

\subsubsection{Deletion Rate}
\begin{figure}
    \centering
    \includegraphics[width=1\linewidth]{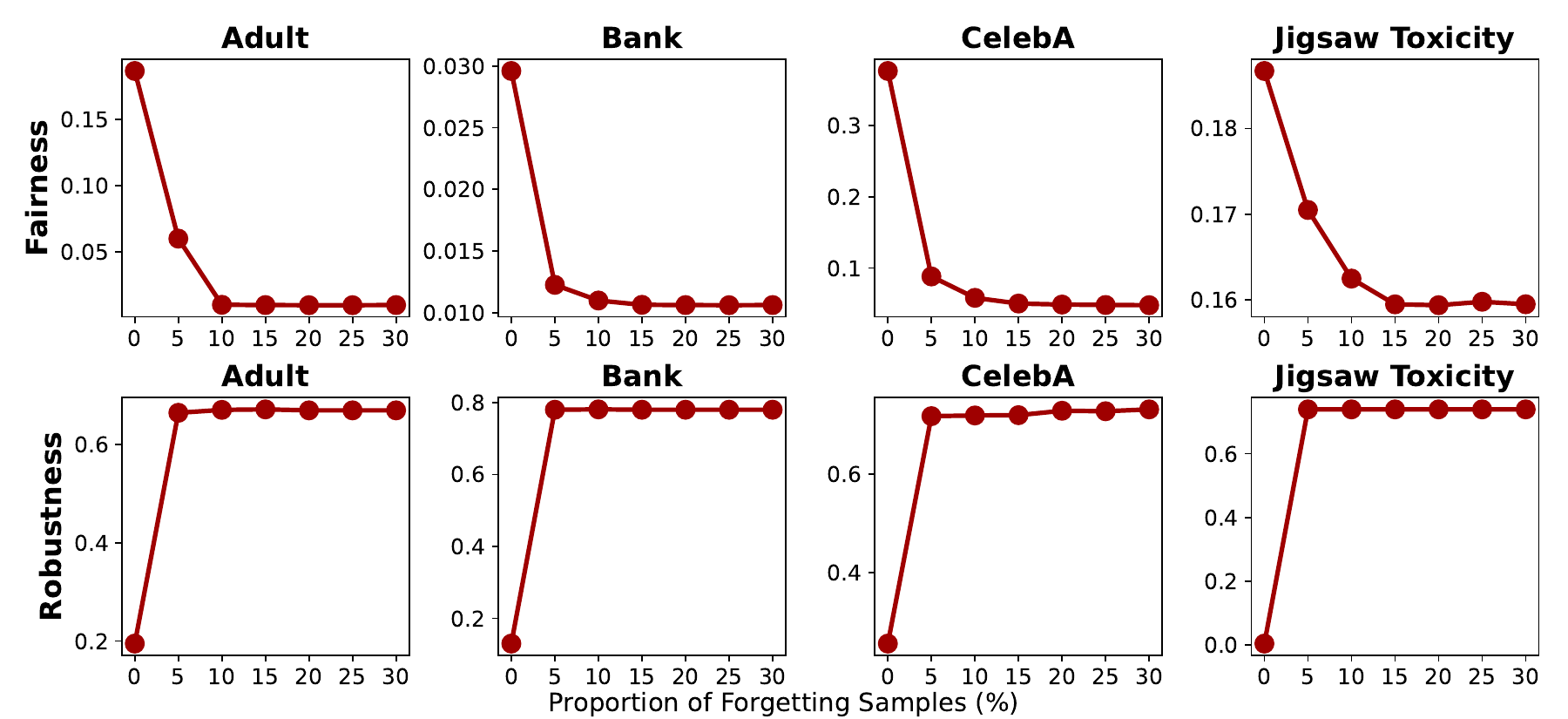}
    \caption{Effect of the proportion of forgetting samples removed on fairness and robustness under the hard-weighted \texttt{(IF)} scheme. As the deletion rate increases from 0\% to 20\%, fairness and robustness metrics improve and then stabilize. Further removal beyond this range yields minimal gains in fairness and robustness. Similar trends are observed for other unlearning methods.}
    \label{Fig: Appendix-4}
\end{figure}
\textbf{Second}, \cref{Fig: Appendix-4} illustrates the process of selecting the number of forgetting samples to remove under the hard-weighted \texttt{(\textbf{IF})} scheme; similar patterns are observed for other methods. As the proportion of removed forgetting samples increases, both fairness and robustness improve accordingly. However, these improvements tend to plateau when the deletion rate reaches approximately 5\% to 20\%. Beyond this range, further increasing the number of removed samples yields diminishing returns in fairness and robustness, while continuing to degrade model accuracy.  However, we observe from \cref{Fig: appendix-time-cost} that the time cost of Step 3 is non-negligible. Each execution of Step 3 requires an expensive matrix inversion, which for a model with parameter dimension $d$ typically incurs a computational complexity of $\mathcal{O}(d^3)$. Searching for the optimal forgetting sample ratio results in a multiplicative increase in computational cost. Therefore, selecting a 20\% deletion rate strikes a practical balance, effectively covering most datasets to ensure maximal improvements in fairness and robustness.

\subsubsection{EOP Results}
\begin{figure}[t]
    \centering
       \includegraphics[width=0.4951 \linewidth]{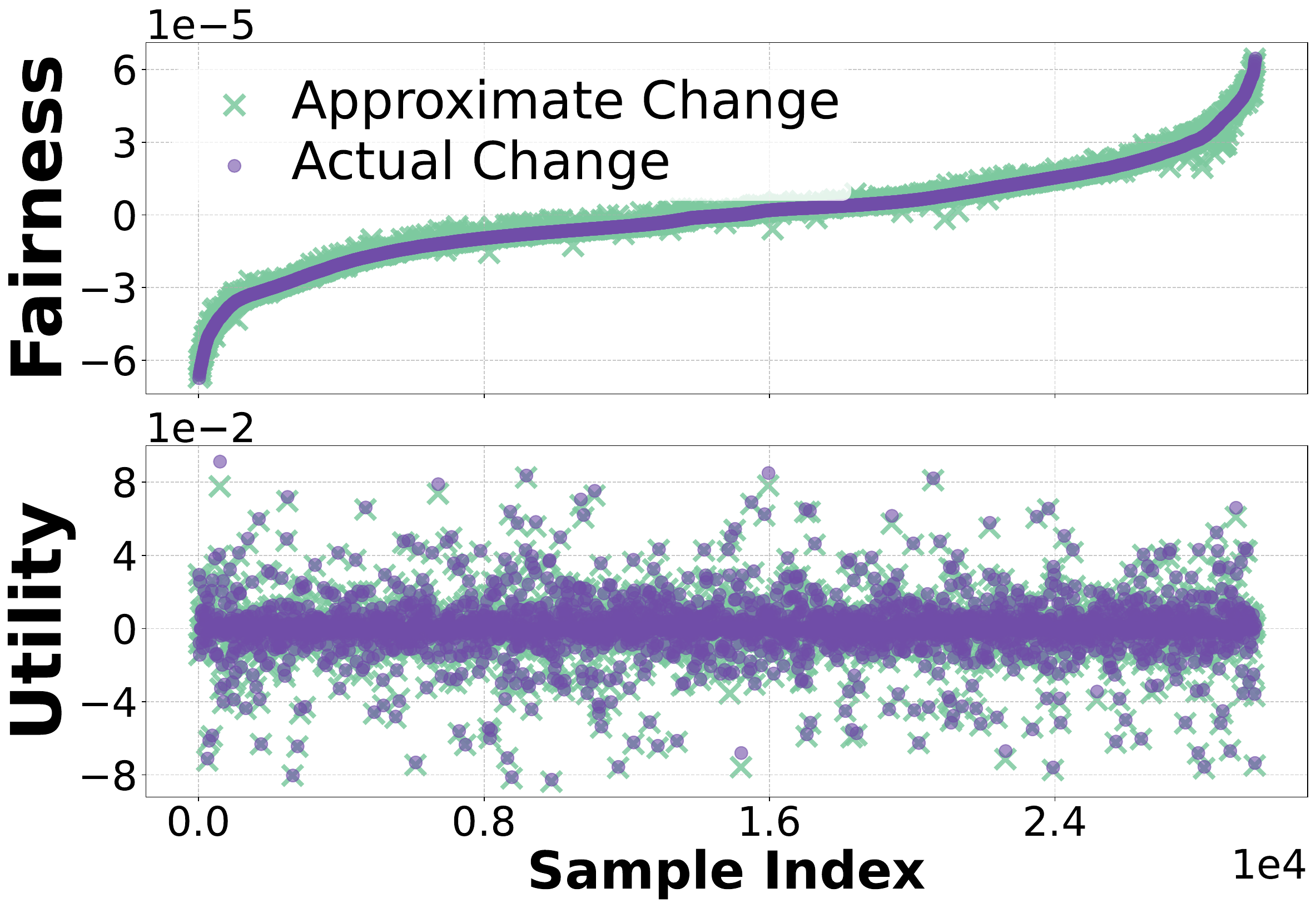}
          \includegraphics[width=0.4951 \linewidth]{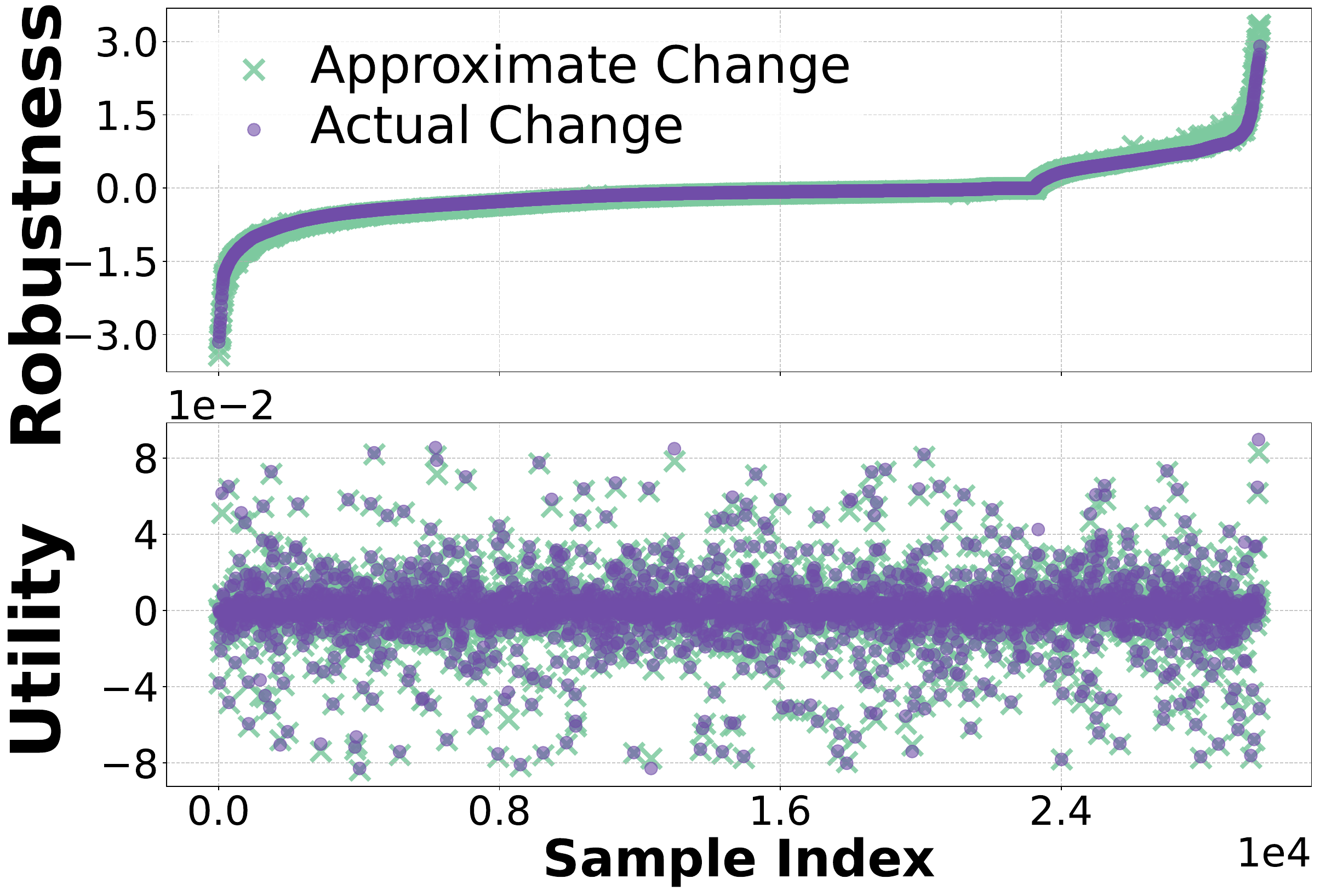}
    \caption{\textbf{Utility Changes vs. Fairness/Robustness Changes.} We evaluated the impact of all training data on different performance metrics as follows:
\textbf{(Left)} The model's generalization ability, evaluated as the loss on the test dataset.
\textbf{(Right)} The model's robustness, evaluated as the loss on adversarial test samples.}
    \label{Fig: SEC-5.2}
\end{figure}

\textbf{Third,} \cref{Fig: SEC-5.2} illustrates the changes in utility across all training samples with respect to both Demographic Parity (DP) and robustness, displaying both the ground-truth values and their approximations. The results suggest that utility is not strongly correlated with either fairness (DP) or robustness, indicating that improvements in these dimensions may not directly translate into gains in utility.

\begin{figure}[t]
    \centering
    \includegraphics[width=0.42\linewidth]{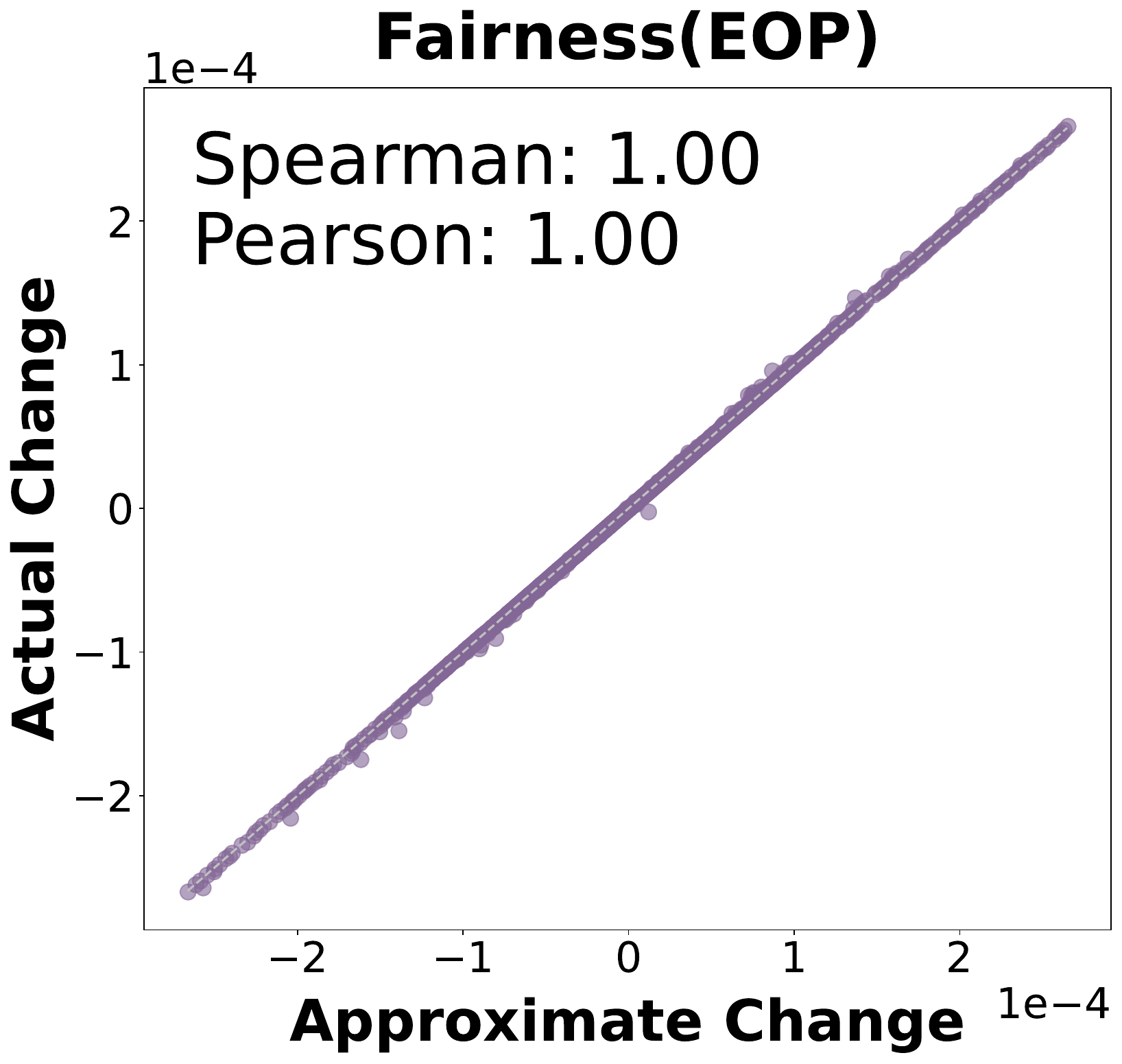}
\includegraphics[width=0.55\linewidth]{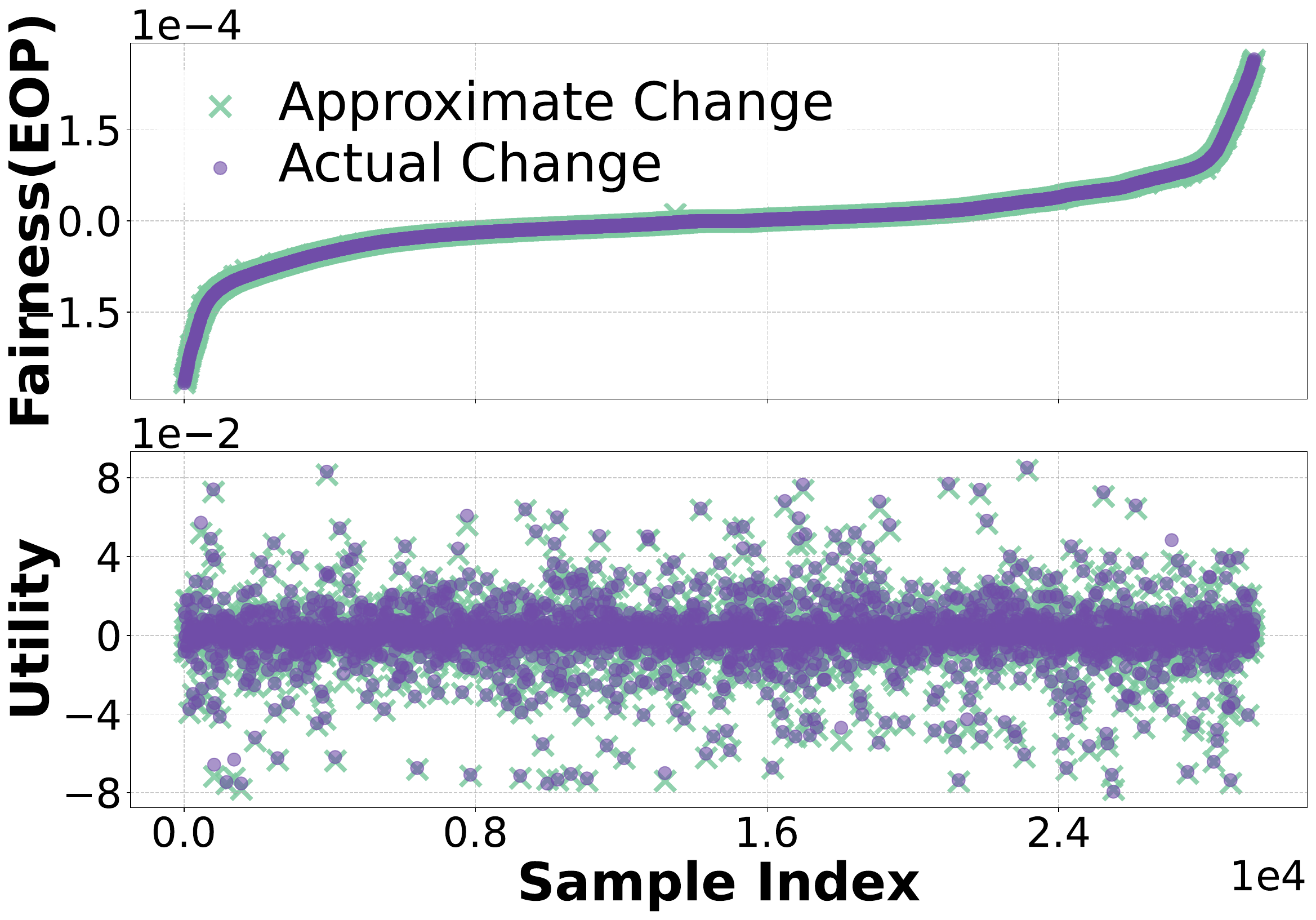}
        \caption{Actual EOP Change vs. Influence EOP Change. The leave-one-out influence of all training samples on the EOP metric. The first plot evaluates the correlation coefficient, indicating an effective approximation of the influence function \textbf{(Left)}. The second plot ranks the samples based on their actual EOP metric from smallest to largest, illustrating the utility of each sample, and suggesting that removing the detrimental samples does not necessarily increase utility \textbf{(Right)}.}
    \label{Fig: Appendix-1}
\end{figure}

\cref{Fig: Appendix-1} further demonstrates that influence functions can accurately approximate the true leave-one-out effects on the model with respect to the Equality of Opportunity (EOP) metric. Importantly, consistent with the observations for DP, removing detrimental samples does not necessarily yield improvements in utility performance.

\begin{figure}[t]
    \centering
\includegraphics[width=0.9\linewidth]{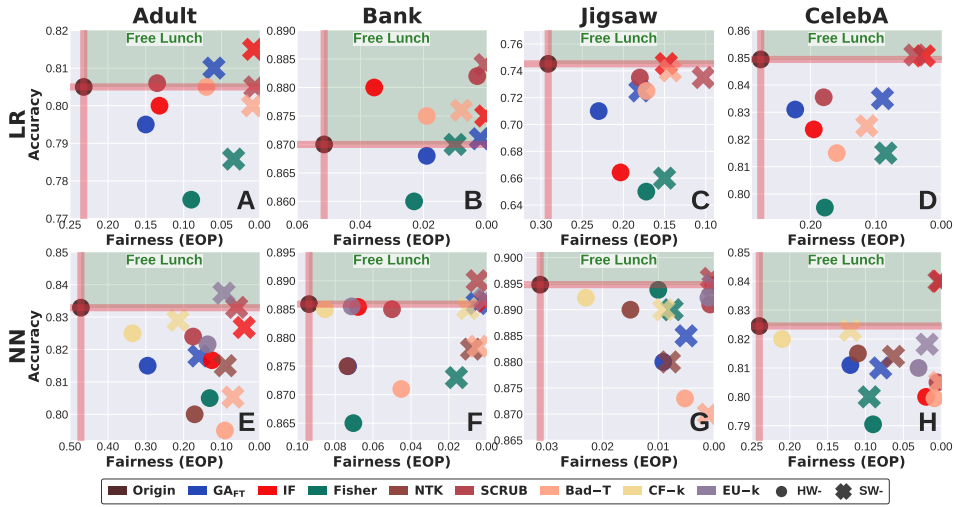}
    \caption{\textbf{Performance on EOP Metric.} Different colors represent various unlearning algorithms: \ding{108} for the hard-weighted scheme and \ding{54} for the soft-weighted scheme. \textbf{The First Two Rows} (LR, NN) evaluate utility and fairness metrics, while \textbf{The Last Two Rows} (LR, NN) evaluate utility and robustness metrics across datasets. \textbf{The Green Region} highlights \textcolor[HTML]{006400}{\textbf{Free Lunch}} cases where unlearning algorithms improve both target task performance and utility compared to the original model. The soft-weighted scheme outperforms the hard-weighted scheme by enhancing task performance and utility, even achieving free lunch in part of unlearning algorithms' results.}
    \label{Fig: Appendix-2}
\end{figure}

\cref{Fig: Appendix-2} presents a comprehensive set of additional experimental results centered on the Equality of Opportunity (EOP) metric. In these experiments, the hard-weighted framework consistently removes 20\% of the training samples. The results clearly demonstrate the advantages of the proposed soft-weighted machine unlearning framework over conventional hard-weighted approaches across a variety of tasks and datasets. Collectively, these findings underscore the framework’s strong potential to address key challenges in machine unlearning, positioning it as a promising solution for both future research and real-world applications.

\subsubsection{Results on Large-Scale Models and Datasets}

\begin{figure}[t]
    \centering
\includegraphics[width=0.73\linewidth]{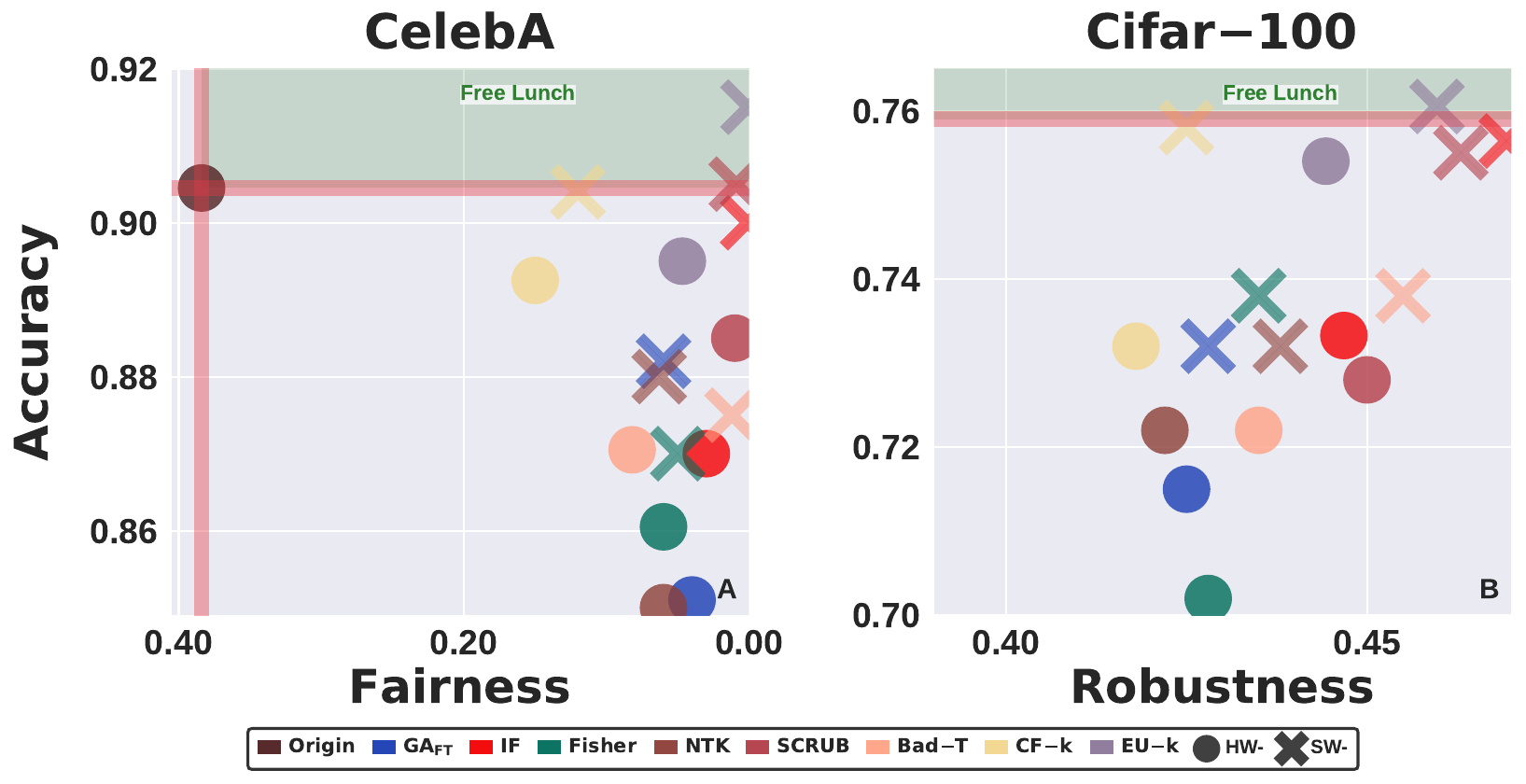}
   \caption{\textbf{Performance on Large-Scale Models and datasets.} \textbf{Left:} ResNet-18 on CelebA for the fairness task. \textbf{Right:} ResNet-50 on CIFAR-100 for the robustness task. \textbf{Green Region:} \textcolor[HTML]{006400}{\textbf{Free Lunch}} cases, where unlearning algorithms improve both the target task performance and overall utility compared to the original model.}
    \label{Fig: Appendix-3}
\end{figure}

\textbf{Fourth}, \cref{Fig: Appendix-3} presents the evaluation results on large-scale settings, including ResNet-18 on CelebA for the fairness task (left) and ResNet-50 on CIFAR-100 for the robustness task (right). We observe consistent trends with smaller models: the proposed soft-weighted unlearning framework achieves competitive or superior performance compared to hard weighted baselines. These results suggest that, even in high-capacity models and more complex datasets, carefully designed unlearning reweighting strategies can enhance both reliability and predictive performance.

\subsubsection{Visualization of the Correlations Between Fairness/Robustness and Utility}
\label{Appendix-Visualizing the Correlations}

\begin{figure}[b]
    \centering
    \includegraphics[width=0.7\linewidth]{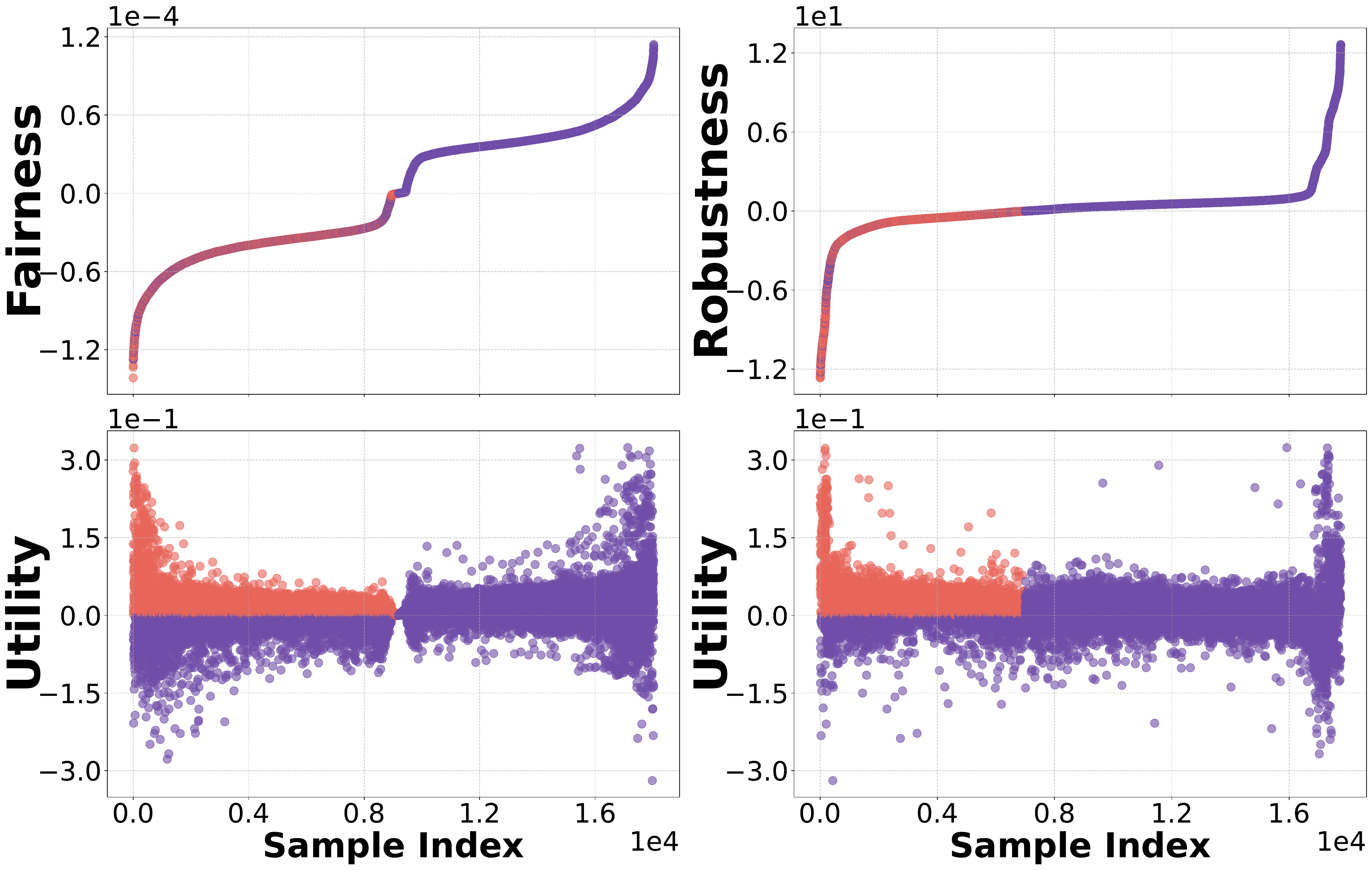}
    \caption{\textbf{Actual Changes in Utility and Fairness/Robustness on Bank} dataset for each sample's leave-one-out model. \textbf{The X-axis represents} the sample indices. \textbf{The Y-axis for Fairness (Robustness)} displays changes in demographic parity (adversarial loss) on the test set, with negative values indicating improved fairness (robustness) and positive values indicating reduced fairness (robustness). \textbf{The Y-axis for Utility} shows changes in test loss, with negative values indicating improved utility. Scatter points marked in \textcolor[HTML]{E8655A}{\textbf{Red}} indicate sample indices where Fairness/Robustness improves, but utility declines.
    }
    \label{Fig: Appendix-3-1}
\end{figure}

\begin{figure}
    \centering
    \includegraphics[width=0.8\linewidth]{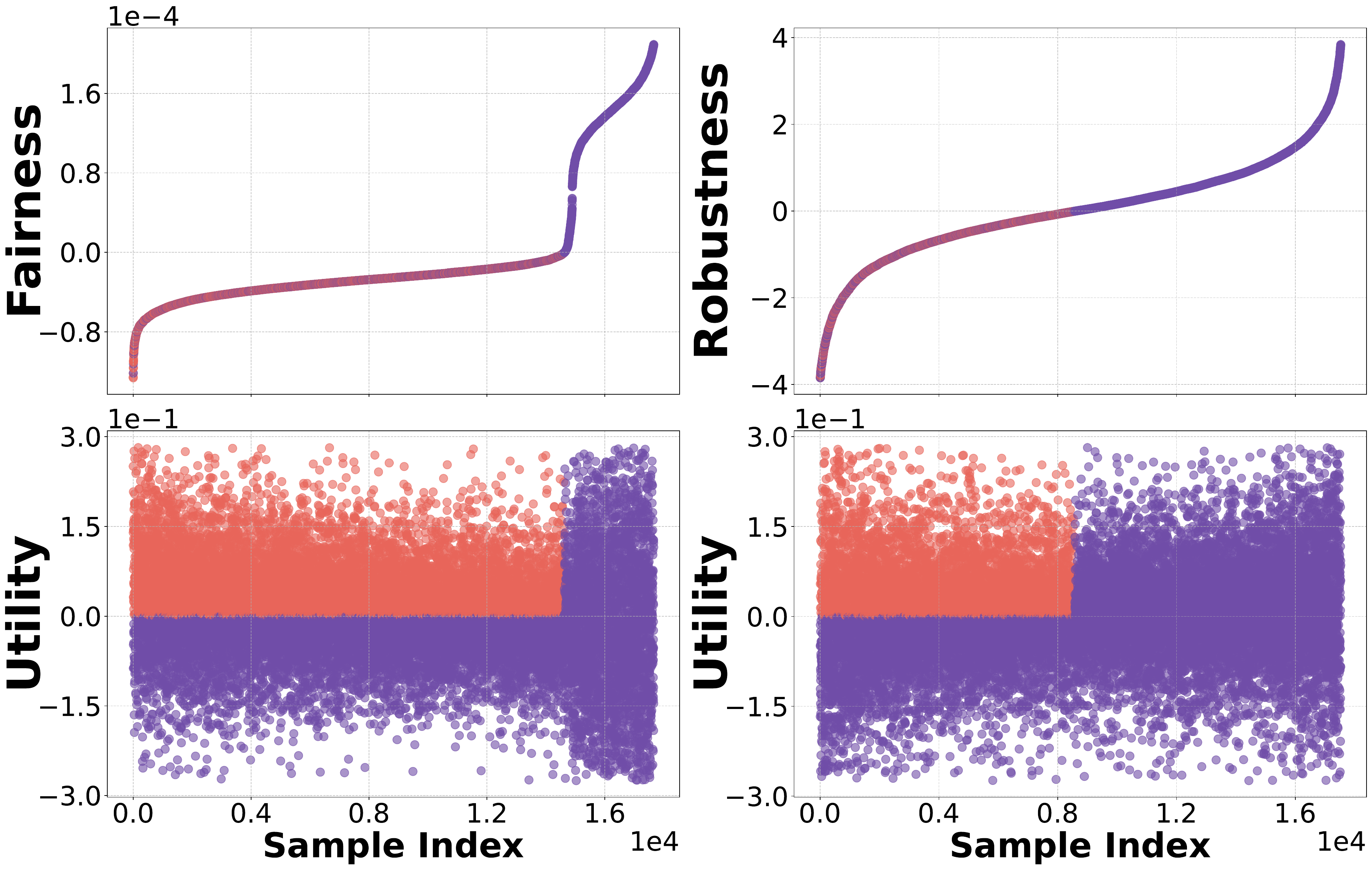}
    \caption{\textbf{Actual Changes in Utility and Fairness/Robustness on Jigsaw dataset} for each sample's leave-one-out model. \textbf{The X-axis represents} the sample indices. \textbf{The Y-axis for Fairness (Robustness)} displays changes in demographic parity (adversarial loss) on the test set, with negative values indicating improved fairness (robustness) and positive values indicating reduced fairness (robustness). \textbf{The Y-axis for Utility} shows changes in test loss, with negative values indicating improved utility. Scatter points marked in \textcolor[HTML]{E8655A}{\textbf{Red}} indicate sample indices where Fairness/Robustness improves, but utility declines.
    }
    \label{Fig: Appendix-3-2}
\end{figure}

\begin{figure}
    \centering
    \includegraphics[width=0.8\linewidth]{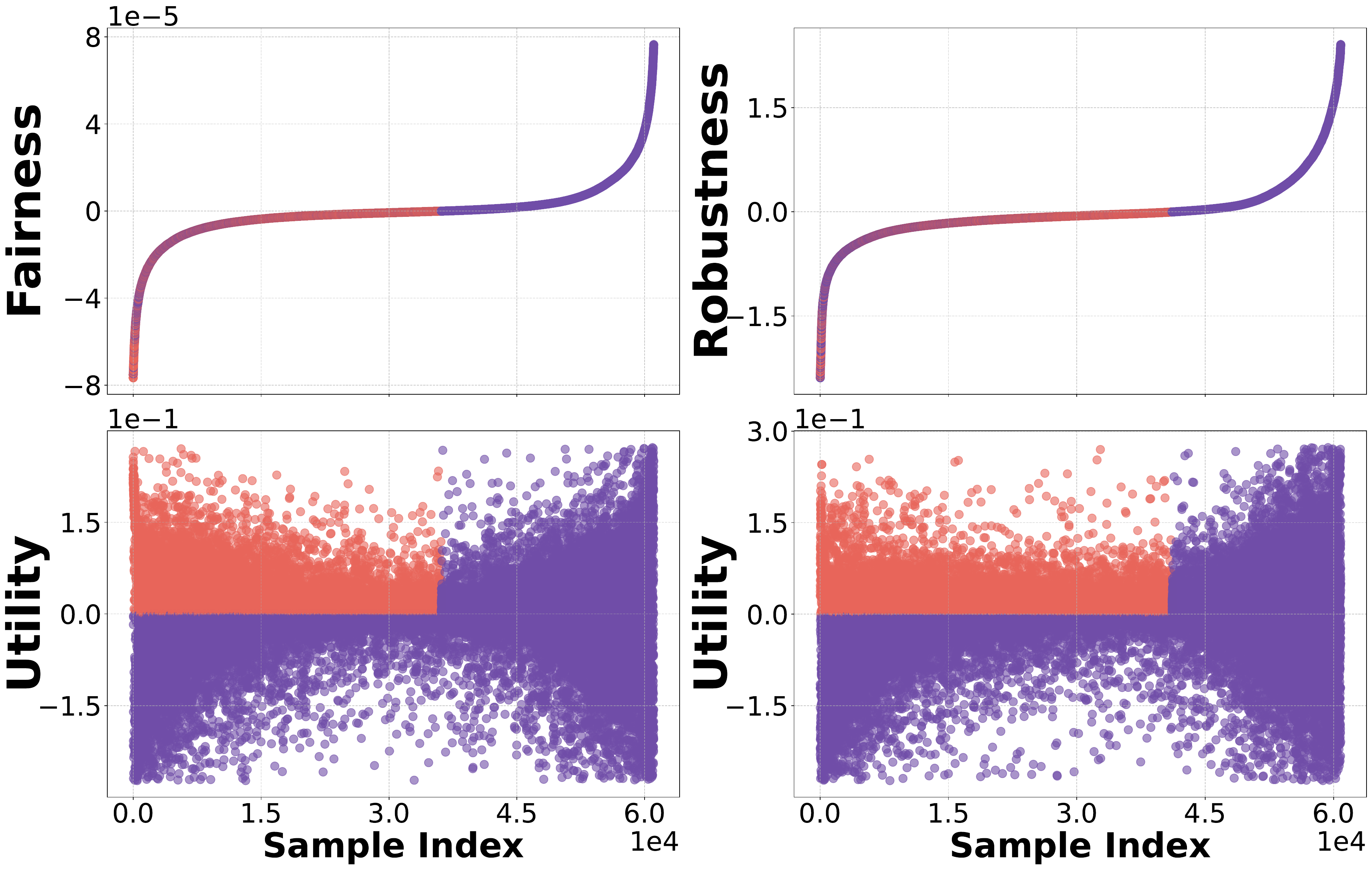}
    \caption{\textbf{Actual Changes in Utility and Fairness/Robustness on CelebA dataset} for each sample's leave-one-out model. \textbf{The X-axis represents} the sample indices. \textbf{The Y-axis for Fairness (Robustness)} displays changes in demographic parity (adversarial loss) on the test set, with negative values indicating improved fairness (robustness) and positive values indicating reduced fairness (robustness). \textbf{The Y-axis for Utility} shows changes in test loss, with negative values indicating improved utility. Scatter points marked in \textcolor[HTML]{E8655A}{\textbf{Red}} indicate sample indices where Fairness/Robustness improves, but utility declines.
    }
    \label{Fig: Appendix-3-3}
\end{figure}

\textbf{Finally}, \cref{Fig: Appendix-3-1,Fig: Appendix-3-2,Fig: Appendix-3-3} show results on additional datasets, which exhibit similar patterns to those observed in \cref{SEC-1}. We trained a linear model on the Bank, CelebA, Jigsaw datasets and analyzed the performance of leave-one-out models obtained by individually removing each training sample. Specifically, we evaluated changes in the following metrics, defined as the differences between their post-removal and pre-removal values: fairness, measured by Demographic Parity \cite{DBLP:conf/innovations/DworkHPRZ12}; adversarial robustness, assessed via the loss on perturbed datasets \cite{DBLP:conf/esann/MegyeriHJ19}; and generalization utility, determined by the loss on the test dataset.

\end{document}